%% file: arxiv.tex
\documentclass[10pt,twocolumn,letterpaper]{article}

\newcommand{\dataurl}{\urlstyle{same}\url{http://cs.nyu.edu/~sermanet/data.html\#inria}}

\newcommand{\figdir}{}
\newcommand{\alldir}{}

\newcommand{\inriadir}{}

\usepackage{cvpr}

\usepackage{times}
\usepackage{epsfig}
\usepackage{graphicx}
\usepackage{amsmath}
\usepackage{amssymb}
\usepackage{algorithm}
\usepackage{algorithmic}
\usepackage{subfigure}
\usepackage{caption}
\usepackage{lscape}
\usepackage{moresize}
\usepackage{anyfontsize}

\newcommand{\inp}{x}
\newcommand{\dic}{{\cal D}}
\newcommand{\code}{z}
\newcommand{\en}{{\mathbb E}}

\newcommand{\Real}{\mathbb{R}}

\newcommand{\argmin}{\arg \min}

\newcommand{\deri}[2]{\frac{\partial  #1}{\partial #2}}
\newcommand{\conv}{\otimes}

\usepackage[pagebackref=true,breaklinks=true,letterpaper=true,colorlinks,bookmarks=false]{hyperref}

\def\cvprPaperID{1129} 

\cvprfinalcopy
\ifcvprfinal\fi

\begin{document}

\title{Pedestrian Detection with\\Unsupervised Multi-Stage Feature Learning}
\author{Pierre Sermanet \and Koray Kavukcuoglu \and Soumith Chintala \and Yann LeCun\\
  Courant Institute of Mathematical Sciences, New York University\\
  {\tt\small {sermanet,koray,soumith,yann}@cs.nyu.edu}
}
\maketitle

\begin{abstract}
Pedestrian detection is a problem of considerable practical interest. Adding to the list of successful applications of deep learning methods to vision, we report state-of-the-art and competitive results on all major pedestrian datasets with a convolutional network model. The model uses a few new twists, such as multi-stage features, connections that skip layers to integrate global shape information with local distinctive motif information, and an unsupervised method based on convolutional sparse coding to pre-train the filters at each stage.
\end{abstract}

\input{section_introduction.tex}
\input{section_algorithm.tex}
\input{section_architecture.tex}

\input{section_experiments.tex}

\input{section_conclusion.tex}

\vspace{-.1in}
\bibliographystyle{latex12}
{\tiny{
  \bibliography{pedestrian}}
}

\onecolumn

\input{supplementary.tex}

\end{document}

%% file: section_introduction.tex
\section{Introduction}

Pedestrian detection is a key problem for surveillance, automotive
safety and robotics applications.  The wide variety of appearances of
pedestrians due to body pose, occlusions, clothing, lighting and
backgrounds makes this task challenging.


All existing state-of-the-art methods use a combination of
hand-crafted features such as \emph{Integral Channel
  Features}~\cite{dollar-bmvc-09}, HoG~\cite{dalal-iccv-05} and their
variations~\cite{felzen-pami-10,5459205} and
combinations~\cite{walk-cvpr-10}, followed by a trainable classifier
such as SVM~\cite{felzen-pami-10,10.1109}, boosted
classifiers~\cite{dollar-bmvc-09} or random forests~\cite{dollarcrosstalk}.
While low-level features can be
designed by hand with good success, mid-level features that combine
low-level features are difficult to engineer without the help of some
sort of learning procedure. Multi-stage recognizers that learn
hierarchies of features tuned to the task at hand can be trained
end-to-end with little prior knowledge.  Convolutional Networks
(ConvNets)~\cite{lecun-98b} are examples of such hierarchical systems
with end-to-end feature learning that are trained in a supervised
fashion.  Recent works have demonstrated the usefulness of
unsupervised pre-training for end-to-end training of deep multi-stage
architectures using a variety of techniques such as stacked restricted
Boltzmann machines~\cite{hinton-science-06}, stacked
auto-encoders~\cite{bengio-nips-07} and stacked sparse
auto-encoders~\cite{ranzato-nips-07}, and using new types of
non-linear transforms at each layer~\cite{jarrett-iccv-09,
  koray-nips-10}.

\begin{figure}[htb]
  \begin{center}
    \includegraphics[width=1\linewidth]{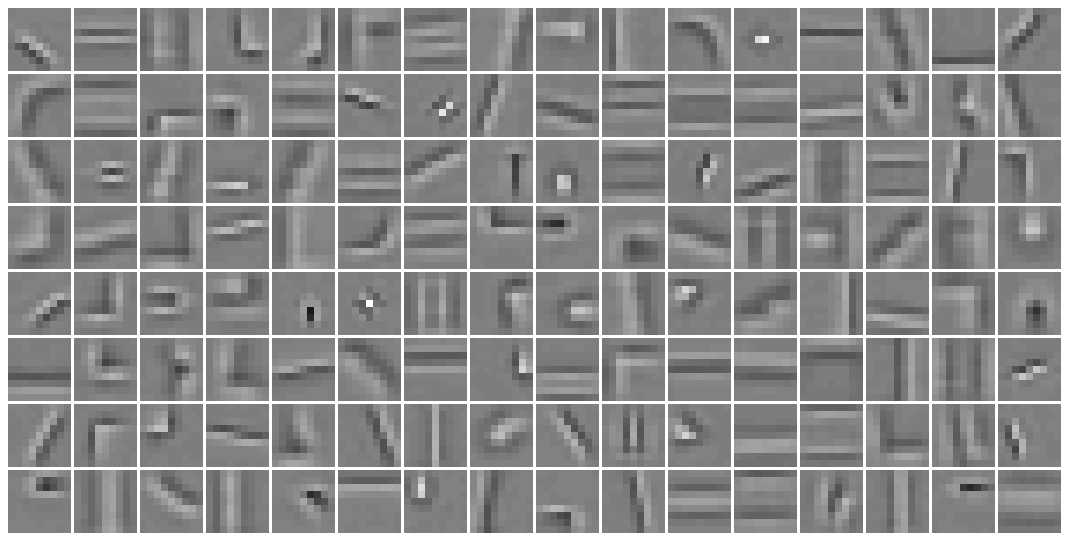}
  \end{center}
  \caption{\small{\textbf{$128$ $9 \times 9$ filters trained on grayscale INRIA
    images using Algorithm~\ref{alg:convpsd}}. It can be seen that in
    addition to edge detectors at multiple orientations, our systems
    also learns more complicated features such as corner and junction
    detectors.}}
  \label{fig:decoder_gray}
\vspace{-0.4cm}
\end{figure}

Supervised ConvNets have been used by a number of authors for such
applications as face, hand
detection~\cite{vaillant-monrocq-lecun-94, nowlan-platt-95,
  garcia-delakis-04, osadchy-07, frome-iccv-09, taylor-nips-10}.
More recently, a large ConvNet by~\cite{kriz12} achieved a breakthrough on a 1000-class ImageNet detection task.
The main contribution of this paper is to show that the ConvNet model,
with a few important twists, consistently yields state of the art and
competitive results
on all major pedestrian detection benchmarks.
The system uses
unsupervised convolutional sparse auto-encoders to pre-train features
at all levels from the relatively small INRIA dataset~\cite{dalal-iccv-05}, and end-to-end supervised training to train the
classifier and fine-tune the features in an integrated fashion.
Additionally, multi-stage features with layer-skipping connections
enable output stages to combine global shape detectors with
local motif detectors.

Processing speed in pedestrian detection has recently seen great progress, enabling real-time operation without sacrificing quality.~\cite{benenson2012pedestrian} manage to entirely avoid image rescaling for detection while observing quality improvements.
While processing speed is not the focus of this paper, features and classifier approximations introduced by~\cite{dollar-bmvc-10}
and~\cite{benenson2012pedestrian} may be applicable to deep learning models for faster detection, in addition to GPU optimizations.

%% file: section_algorithm.tex
\section{Learning Feature Hierarchies}

Much of the work on pedestrian detection have focused on designing
representative and powerful features~\cite{dalal-iccv-05,
  dollar-bmvc-09, dollar-bmvc-10, walk-cvpr-10}. In this work, we show
that generic feature learning algorithms can produce successful
feature extractors that can achieve state-of-the-art results.

Supervised learning of end-to-end systems on images have been shown to
work well when there is abundant labeled samples~\cite{lecun-98b},
including for detection tasks~\cite{vaillant-monrocq-lecun-94,
  nowlan-platt-95, garcia-delakis-04, osadchy-07, frome-iccv-09,
  taylor-nips-10}. However, for many input domains, it is hard to find
adequate number of labeled data. In this case, one can resort to
designing useful features by using domain knowledge, or an alternative
way is to use unsupervised learning algorithms. Recently unsupervised
learning algorithms have been demonstrated to produce good features
for generic object recognition problems~\cite{lee-nips-07,
  lee-icml-09, koray-cvpr-09, koray-nips-10}.

In~\cite{hinton-science-06}, it was shown that unsupervised learning
can be used to train deep hierarchical models and the final
representation achieved is actually useful for a variety of different
tasks~\cite{ranzato-nips-07, lee-nips-07, bengio-nips-07}. In this
work, we also follow a similar approach and train a generic
unsupervised model at each layer using the output representation from
the layer before. This process is then followed by supervised updates
to the whole hierarchical system using label information.

\begin{figure}[htb]
  \begin{center}
    \includegraphics[width=1\linewidth]{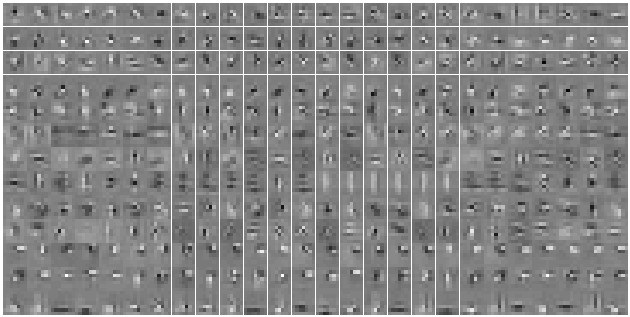}
  \end{center}
  \caption{\small{\textbf{A subset of $7 \times 7$ second layer filters} trained on
    grayscale INRIA images using Algorithm~\ref{alg:mconvpsd}. Each
    row in the figure shows filters that connect to a common output
    feature map. It can be seen that they extract features at similar
    locations and shapes, e.g. the bottom row tends to aggregate horizontal
  features towards the bottom of the filters.}}
  \label{fig:decoder_gray2}
\end{figure}

\subsection{Hierarchical Model}

A hierarchical feature extraction system consists of multiple levels
of feature extractors that perform the same filtering and non-linear
transformation functions in successive layers. Using a particular
generic parametrized function one can then map the inputs into
gradually more higher level (or abstract)
representations~\cite{lecun-98b, hinton-science-06, bengio-nips-07,
  ranzato-nips-07, lee-nips-07}. In this work we use sparse
convolutional feature hierarchies as proposed
in~\cite{koray-nips-10}. Each layer of the unsupervised model contains
a convolutional sparse coding algorithm and a predictor function that
can be used for fast inference. After the last layer a classifier is
used to map the feature representations into class labels. Both the
sparse coding dictionary and the predictor function do not contain any
hard-coded parameter and are trained from the input data.

The training procedure for this model is similar
to~\cite{hinton-science-06}. Each layer is trained in an unsupervised
manner using the representation from previous layer (or the input
image for the initial layer) separately. After the whole multi-stage
system is trained in a layer-wise fashion, the complete architecture
followed by a classifier is fine-tuned using labeled data.

\subsection{Unsupervised Learning}

Recently sparse coding has seen much interest in many fields due to
its ability to extract useful feature representations from data, The
general formulation of sparse coding is a linear reconstruction model
using an overcomplete dictionary $\dic \in \Real^{m \times n}$ where
$m > n$ and a regularization penalty on the mixing coefficients $\code
\in \Real^{n}$.
\begin{align}
  \code^* = \argmin_{\code} \| \inp - \dic \code \|_2^2 + \lambda s(\code)
  \label{eqn:sc}
\end{align}
The aim is to minimize equation~\ref{eqn:sc} with respect to $\code$
to obtain the optimal sparse representation $\code^*$ that correspond
to input $\inp \in \Real^{m}$. The exact form of $s(\code)$ depends on
the particular sparse coding algorithm that is used, here, we use the
$\|.\|_1$ norm penalty, which is the sum of the absolute values of all
elements of $z$. It is immediately clear that the solution of this
system requires an optimization process. Many efficient algorithms for
solving the above convex system has been proposed in recent
years~\cite{aharon-ksvd-05, daubechies-ista-2004, beck-fista-09, li-osher-10}. However, our aim is to also learn generic feature
extractors. For that reason we minimize equation~\ref{eqn:sc} wrt
$\dic$ too.
\begin{align}
  \code^*,\dic^* = \argmin_{\code,\dic} \| \inp - \dic \code \|_2^2 + \lambda \|z\|_1
  \label{eqn:sm}
\end{align}
This resulting equation is non-convex in $\dic$ and $\code$ at the
same time, however keeping one fixed, the problem is still convex wrt
to the other variable. All sparse modeling algorithms that adopt the
dictionary matrix $\dic$ exploit this property and perform a
coordinate descent like minimization process where each variable is
updated in succession. Following~\cite{olshausen-97} many authors have
used sparse dictionary learning to represent
images~\cite{mairal-cvpr-08, aharon-ksvd-05, koray-psd-08}. However,
most of the sparse coding models use small image patches as input
$\inp$ to learn the dictionary $\dic$ and then apply the resulting
model to every overlapping patch location on the full image. This
approach assumes that the sparse representation for two neighboring
patches with a single pixel shift is completely independent, thus
produces very redundant representations.~\cite{koray-nips-10,
  zeiler-cvpr-10} have introduced convolutional sparse modeling
formulations for feature learning and object recognition and we use
the Convolutional Predictive Sparse Decomposition (CPSD) model
proposed in~\cite{koray-nips-10} since it is the only convolutional
sparse coding model providing a fast predictor function that is
suitable for building multi-stage feature representations. The
particular predictor function we use is similar to a single layer
ConvNet of the following form:
\begin{align}
  f(\inp;g,k,b) & = \tilde{\code} = \{\tilde{\code}_j\}_{j=1..n}\\
  \tilde{\code}_j & = g_j \times \tanh(\inp \conv k_j + b_j)
\end{align}
where $\conv$ operator represents convolution operator that applies on
a single input and single filter. In this formulation $\inp$ is a $p
\times p$ grayscale input image, $k \in \Real^{n \times m \times m}$
is a set of 2D filters where each filter is $k_j \in \Real^{m \times
  m}$, $g \in \Real^n$ and $b \in \Real^n$ are vectors with $n$
elements, the predictor output $\tilde{z} \in \Real^{n \times p-m+1
  \times p-m+1}$ is a set of feature maps where each of $\tilde{z}_j$
is of size $p-m+1 \times p-m+1$. Considering this general predictor
function, the final form of the convolutional unsupervised energy for
grayscale inputs is as follows:

\begin{align}
  \en_{CPSD} & = \en_{ConvSC} + \beta \en_{Pred} \label{eqn:cpsd}\\
  \en_{ConvSC} & = \left \| \inp - {\sum}_j \dic_j \conv \code_j \right \|_2^2  + \lambda \|z\|_1 \label{eqn:csc}\\
  \en_{Pred} & = \| \code^* - f(\inp;g,k,b) \|_2^2 \label{eqn:pred}
\end{align}
where $\dic$ is a dictionary of filters the same size as $k$ and
$\beta$ is a hyper-parameter. The unsupervised learning procedure is a
two step coordinate descent process. At each iteration, {\bf (1)
  Inference:} The parameters $W = \{\dic, g, k, b\}$ are kept fixed
and equation~\ref{eqn:csc} is minimized to obtain the optimal sparse
representation $\code^*$, {\bf (2) Update:} Keeping $\code^*$ fixed,
the parameters $W$ updated using a stochastic gradient step: $ W
\leftarrow W - \eta \deri{\en_{CPSD}}{W}$ where $\eta$ is the learning
rate parameter. The inference procedure requires us to carry out the
sparse coding problem solution. For this step we use the FISTA method
proposed in~\cite{beck-fista-09}. This method is an extension of the
original iterative shrinkage and thresholding
algorithm~\cite{daubechies-ista-2004} using an improved step size
calculation with a momentum-like term. We apply the FISTA algorithm in
the image domain adopting the convolutional formulation.

For color images or other multi-modal feature representations, the
input $\inp$ is a set of feature maps indexed by $i$ and the
representation $\code$ is a set of feature maps indexed by $j$ for
each input map $i$. We define a map of connections $P$ from input
$\inp$ to features $\code$. A $j^{th}$ output feature map is connected
to a set $P_j$ of input feature maps. Thus, the predictor function in
Algorithm~\ref{alg:convpsd} is defined as:
\begin{align}
  \tilde{z}_j = g_j \times \tanh{\left( \sum_{i \in P_j} \left( \inp_i \conv k_{j,i} \right) + b_j \right)}
  \label{eqn:convpred}
\end{align}
and the reconstruction is computed using the inverse map $\bar{P}$:
\begin{align}
  \en_{ConvSC} = \sum_i \| \inp_i - \sum_{j \in \bar{P_i}}\dic_{i,j} \conv \code_j \|_2^2 + \lambda \|z\|_1
  \label{eqn:convsc}
\end{align}
For a fully connected layer, all the input features are connected to
all the output features, however it is also common to use sparse
connection maps to reduce the number of parameters. The online
training algorithm for unsupervised training of a single layer is:
\begin{algorithm}
  \caption{Single layer unsupervised training.}
  \label{alg:convpsd}
\begin{algorithmic}
  \FUNCTION{${\rm\bf Unsup}(\inp, \dic, P, \{ \lambda, \beta \} , \{ g, k, b \},\eta)$}
    \STATE {\bfseries Set:} $f(\inp;g,k,b)$ from eqn~\ref{eqn:convpred}, $W^p = \{ g,k,b \}$.
    \STATE {\bfseries Initialize:} $\code=\emptyset$, $\dic$ and $W^p$ randomly.
    \REPEAT
    \STATE Perform {\bf inference}, minimize equation~\ref{eqn:convsc} wrt $\code$ using FISTA~\cite{beck-fista-09}
    \STATE Do a stochastic {\bf update} on $\dic$ and $W^p$. $\dic \leftarrow \dic - \eta \deri{\en_{ConvSC}}{\dic}$ and $W^p \leftarrow W^p - \eta \deri{\en_{Pred}}{W^p}$
    \UNTIL{convergence}
    \STATE {\bfseries Return:} $\{\dic,g,k,b \}$
  \ENDFUNCTION
\end{algorithmic}
\end{algorithm}

\vspace{-0.5cm}

\subsection{Non-Linear Transformations}
Once the unsupervised learning for a single stage is completed, the
next stage is trained on the feature representation from the previous
one. In order to obtain the feature representation for the next stage,
we use the predictor function $f(x)$ followed by non-linear
transformations and pooling. Following the multi-stage framework used
in~\cite{koray-nips-10}, we apply absolute value rectification, local
contrast normalization and average down-sampling operations.

\noindent
{\bf Absolute Value Rectification} is applied component-wise
to the whole feature output from $f(x)$ in order to avoid cancellation
problems in contrast normalization and pooling steps.  

\noindent
{\bf Local Contrast Normalization} is a non-linear process that enhances
the most active feature and suppresses the other ones. The exact form
of the operation is as follows:
\begin{align}
      v_{i} & = x_{i} - x_i \conv w\;,\;\;\;\sigma = \sqrt{\sum_{i} w \conv v_i^2} \label{eqn:lcn1}\\
      y_{i} & = \frac{v_{i}}{{\rm max}(c,\sigma)}  \label{eqn:lcn3}
\end{align}
where $i$ is the feature map index and $w$ is a $9 \times 9$ Gaussian
weighting function with normalized weights so that $\sum_{ipq}
w_{pq}=1$. For each sample, the constant $c$ is set to $mean(\sigma)$
in the experiments.

\noindent
{\bf Average Down-Sampling} operation is performed using a fixed size
boxcar kernel with a certain step size. The size of the kernel and the
stride are given for each experiment in the following sections.

Once a single layer of the network is trained, the features for
training a successive layer is extracted using the predictor function
followed by non-linear transformations. Detailed procedure of training
an $N$ layer hierarchical model is explained in
Algorithm~\ref{alg:mconvpsd}.

\begin{algorithm}
  \caption{Multi-layer unsupervised training.}
  \label{alg:mconvpsd}
\begin{algorithmic}
  \FUNCTION{${\rm\bf HierarUnsup}(\inp, n_i, m_i, P_i, \{ \lambda_i, \beta_i \}, \{ w_i, s_i \},$\\ $i = \{1 .. N\}, \eta_i)$}
  \STATE {\bfseries Set:} $i=1$, $X_1 = \inp$, $lcn(x)$ using equations~\ref{eqn:lcn1}-\ref{eqn:lcn3}, $ds(X,w,s)$ as the down-sampling operator using boxcar kernel of size $w \times w$ and stride of size $s$ in both directions.
  \REPEAT
   \STATE {\bfseries Set:}  $\dic_i,k_i \in \Real^{n_i \times m_i \times m_i}$, $g_i,b_i \in \Real^{n_i}$.
   \STATE $\{\dic_i,k_i,g_i,k_i,b_i\} = $\\ $\;\;\;\;\;\;\;Unsup(X_i, \dic_i, P_i, \{\lambda_i,\beta_i\}, \{g_i,k_i,b_i\},\eta_i)$
   \STATE $\tilde{z} = f(X_i;g_i,k_i,b_i)$ using equation~\ref{eqn:convpred}.
   \STATE $\tilde{z} = |\tilde{z}|$
   \STATE $\tilde{z} = lcn(\tilde{z})$
   \STATE $X_{i+1} = ds(\tilde{z},w_i, s_i)$
   \STATE $i = i+1$
  \UNTIL{$i=N$}
  \ENDFUNCTION
\end{algorithmic}
\end{algorithm}

The first layer features can be easily displayed in the parameter
space since the parameter space and the input space is same, however
visualizing the second and higher level features in the input space
can only be possible when only invertible operations are used in
between layers. However, since we use absolute value rectification and
local contrast normalization operations mapping the second layer
features onto input space is not possible. In
Figure~\ref{fig:decoder_gray2} we show a subset of $1664$ second layer
features in the parameter space.

\subsection{Supervised Training}
After the unsupervised learning of the hierarchical feature extraction
system is completed using Algorithm~\ref{alg:mconvpsd}, we append a
classifier function, usually in the form of a linear logistic
regression, and perform stochastic online training using labeled data. 

%% file: section_architecture.tex
\subsection{Multi-Stage Features}

\begin{figure}
  \begin{center}
    \includegraphics[width=1\linewidth]{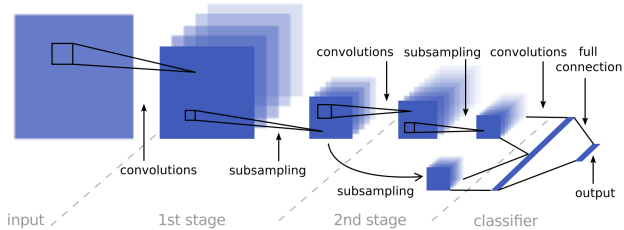}
  \end{center}
  \caption{\textbf{A multi-scale convolutional network.}  The top row of maps
    constitute a regular ConvNet~\cite{jarrett-iccv-09}. The bottom row
    in which the 1st stage output is branched, subsampled again and
    merged into the
    classifier input provides a multi-stage component to the classifier
    stage. The multi-stage features coming out of the 2nd stage
    extracts a global structure as well as local details.}
  \label{fig:net}
\end{figure}

\begin{table*}[htb] 
  \small{
    \begin{center}
      \begin{tabular}{ | c | c | c | c |}
        \hline
        Task & Single-Stage features & Multi-Stage features & Improvement \% \\
        \hline \hline
        Pedestrians detection (INRIA) (Fig.~\ref{fig:inria})& 23.39\% & 17.29\% & 26.1\%\\
        Traffic Signs classification (GTSRB)~\cite{sermanet-ijcnn-11}& 1.80\% & 0.83\% & 54\%\\
        House Numbers classification (SVHN)~\cite{sermanet-icpr-12}& 5.54\% & 5.36\% & 3.2\%\\
        \hline
      \end{tabular}
    \end{center}
    \caption{\small{\textbf{Error rates improvements of multi-stage features over
        single-stage features} for different types of objects detection and
        classification. Improvements are significant for multi-scale and
        textured objects such as traffic signs and pedestrians but minimal for
        house numbers.}} 
    \label{table:ms}
  }
\end{table*}

ConvNets are usually organized in a strictly feed-forward manner where
one layer only takes the output of the previous layer as
input. Features extracted this way tend to be high level features
after a few stages of convolutions and subsampling. By branching lower
levels' outputs into the top classifier (Fig.~\ref{fig:net}), one
produces features that extract both global shapes and structures and
local details, such as a global silhouette and face components in the
case of human detection.  Contrary to~\cite{fan-nn-10}, the output of
the first stage is branched after the non-linear transformations and
pooling/subsampling operations rather than before.

We also use color information on the training data. For this purpose
we convert all images into YUV image space and subsample the UV
features by 3 since color information is in much lower
resolution. Then at the first stage, we keep feature extraction
systems for Y and UV channels separate. On the Y channel, we use $32$
$7 \times 7$ features followed by absolute value rectification,
contrast normalization and $3 \times 3$ subsampling. On the
subsampled UV channels, we extract $6$ $5 \times 5$ features followed
by absolute value rectification and contrast normalization, skipping
the usual subsampling step since it was performed beforehand. These
features are then concatanated to produce $38$ feature maps that are
input to the first layer. The second layer feature extraction takes
$38$ feature maps and produces $68$ output features using $2040$ $9
\times 9$ features. A randomly selected 20\% of the connections in
mapping from input features to output features is removed to limit the
computational requirements and break the
symmetry~\cite{lecun-98b}. The output of the second layer features are
then transformed using absolute value rectification and contrast
normalization followed by $2 \times 2$ subsampling. This results in
$17824$ dimensional feature vector for each sample which is then fed
into a linear classifier.

In Table~\ref{table:ms}, we show that multi-stage features improve accuracy for different tasks, with different magnitudes. Greatest improvements are obtained for pedestrian detection and traffic-sign classification while only minimal gains are obtained for house numbers classification, a less complex task.

\subsection{Bootstrapping}

Bootstrapping is typically used in detection settings by multiple phases
of extracting
the most offending negative answers and adding these samples
to the existing dataset while training. For this purpose, we
extract 3000 negative samples per bootstrapping pass and limit the
number of most offending answers to 5 for each image. We perform 3
bootstrapping passes in addition to the original training phase
(i.e. 4 training passes in total).

\subsection{Non-Maximum Suppression}

Non-maximum suppression (NMS) is used to resolve conflicts when
several bounding boxes overlap. For both INRIA and Caltech experiments
we use the widely accepted PASCAL overlap criteria to determine a
matching score between two bounding boxes
($\frac{intersection}{union}$) and if two boxes overlap by more than
60\%, only the one with the highest score is kept.
In~\cite{dollarCVPR09peds}'s addendum, the matching criteria is
modified by replacing the union of the two boxes with the minimum of
the two. Therefore, if a box is fully contained in another one the
small box is selected. The goal for this modification is to avoid
false positives that are due to pedestrian body parts. However, a
drawback to this approach is that it always disregards one of the
overlapping pedestrians from detection. Instead of changing the
criteria, we actively modify our training set before each
bootstrapping phase. We include body part images that cause false
positive detection into our bootstrapping image set. Our model can
then learn to suppress such responses within a positive window and
still detect pedestrians within bigger windows more reliably.

%% file: section_experiments.tex
\section{Experiments}

We evaluate our system on 5 standard pedestrian detection
datasets. However, like most other systems, we only
train on the INRIA dataset. We also demonstrate improvements brought by unsupervised training and
multi-stage features. In the following we name our model {\bf ConvNet} with variants of unsupervised (Convnet-U) and fully-supervised training (Convnet-F)
and multi-stage features (Convnet-U-MS and ConvNet-F-MS).

\subsection{Data Preparation}

The ConvNet is trained on the INRIA pedestrian
dataset~\cite{dalal-iccv-05}.  Pedestrians are extracted into windows
of 126 pixels in height and 78 pixels in width. The context ratio is
1.4, i.e. pedestrians are 90 pixels high and the remaining 36 pixels
correspond to the background. Each pedestrian image is mirrored along
the horizontal axis to expand the dataset. Similarly, we add 5
variations of each original sample using 5 random deformations such as
translations and scale. Translations range from -2 to 2 pixels and
scale ratios from 0.95 to 1.05.  These deformations enforce invariance
to small deformations in the input. The range of each deformation
determines the trade-off between recognition and localization accuracy
during detection. An equal amount of background samples are extracted
at random from the negative images and taking approximately 10\% of
the extracted samples for validation yields a validation set with 2000
samples and training set with 21845 samples. Note that the unsupervised
training phase is performed on this initial data before the bootstrapping
phase.

\begin{figure*}[htb]

\mbox{\subfigure{\includegraphics[width=.5\linewidth]{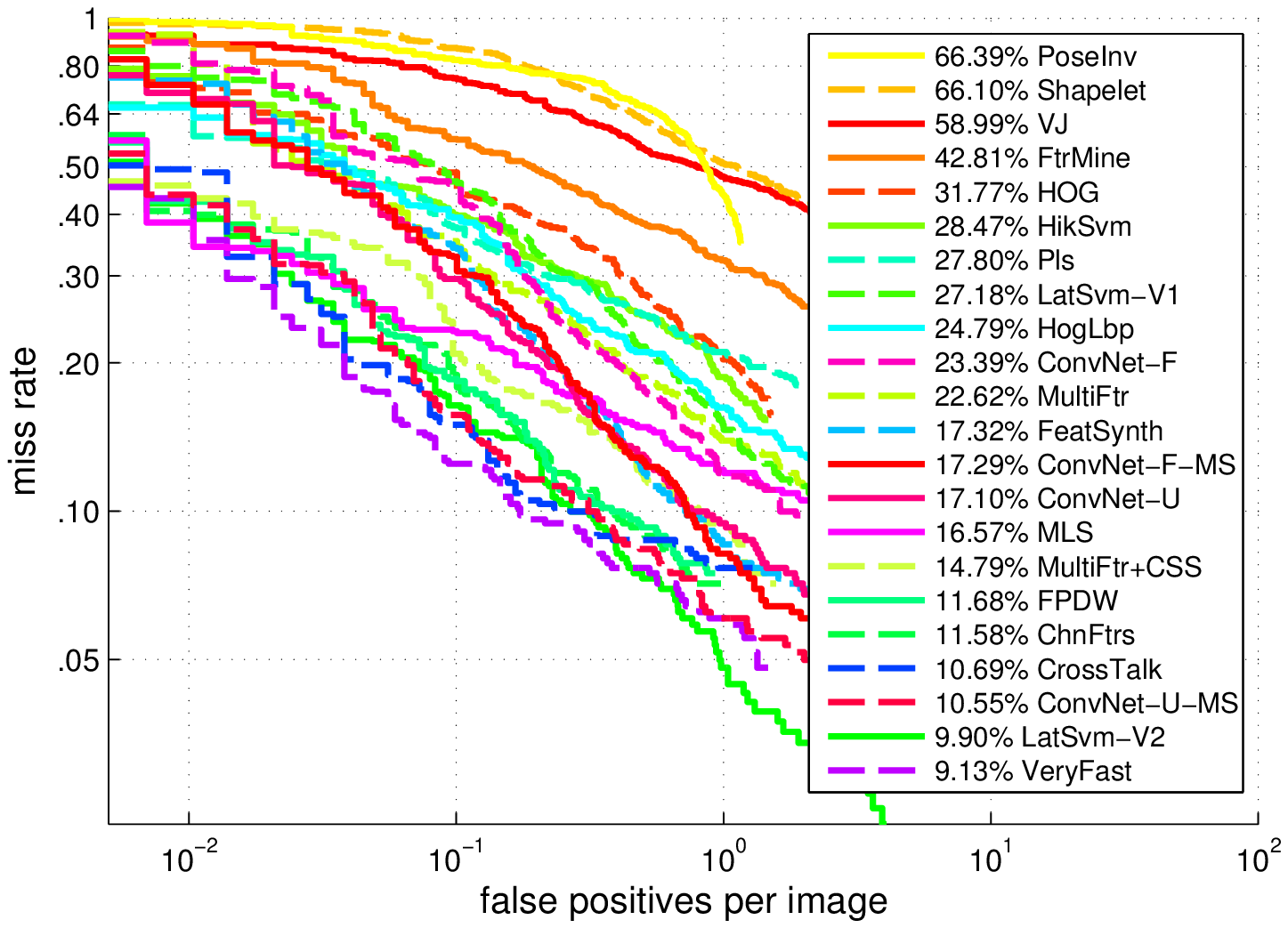}
\quad
\subfigure{\includegraphics[width=.5\linewidth]{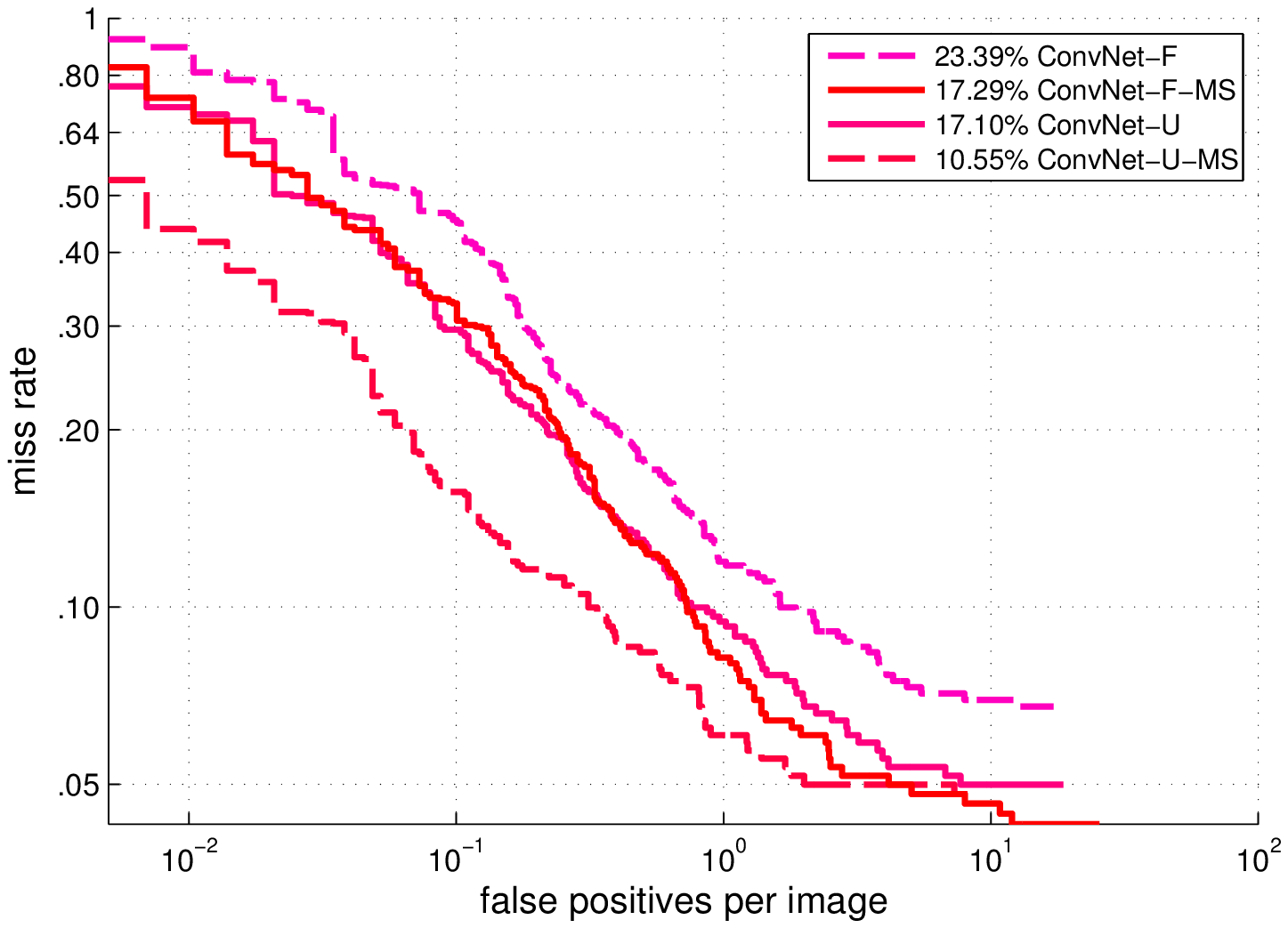} }}}
\vspace{-0.25cm}
  \begin{center}
  \caption{\small{\textbf{DET curves on the fixed-INRIA dataset for large pedestrians measure}
    report false positives per image
    (FPPI) against miss rate. Algorithms are sorted from top to bottom
    using the proposed continuous area under curve measure between 0 and 1 FPPI.
    \textbf{On the right, only the ConvNet variants are displayed to highlight the
    individual contributions of unsupervised
    learning (ConvNet-U) and multi-stage features learning
    (ConvNet-F-MS) and their combination (ConvNet-U-MS)
    compared to the fully-supervised system without multi-stage features
    (ConvNet-F).}}}
  \label{fig:inria}
  \end{center}
\end{figure*}


\subsection{Evaluation Protocol}

During testing and bootstrapping phases using the INRIA dataset, the
images are both up-sampled and sub-sampled. The up-sampling ratio is 1.3
while the sub-sampling ratio is limited by 0.75 times the network's
minimum input ($126 \times 78$). We use a scale stride of 1.10 between each
scale, while other methods typically use either 1.05 or
1.20~\cite{Dollar2011PAMI}.
A higher scale stride is desirable as it implies less computations.

For evaluation we use the bounding boxes files published on the
Caltech Pedestrian
website~\footnote{\urlstyle{same}\url{http://www.vision.caltech.edu/Image\_Datasets/CaltechPedestrians}} and the evaluation software provided by Piotr Dollar (version 3.0.1). In an effort to provide a more accurate evaluation, we improved on both the evaluation formula and the INRIA annotations as follows. The evaluation software was slightly modified to compute the continuous area under curve (AUC) in the entire [0, 1] range rather than from 9 discrete points only (0.01, 0.0178, 0.0316, 0.0562, 0.1, 0.1778, 0.3162, 0.5623 and 1.0 in version 3.0.1). Instead, we compute the entire area under the curve by summing the areas under the piece-wise linear interpolation of the curve, between each pair of points. In addition, we also report a 'fixed' version of the annotations for INRIA dataset, which has missing positive labels. The added labels are only used to avoid counting false errors and wrongly penalizing algorithms. The modified code and extra INRIA labels are available at~\footnote{\dataurl}. Table~\ref{table:all} reports results for both original and fixed INRIA datasets. Notice that the continuous AUC and fixed INRIA annotations both yield a reordering of the results (see supplementary material for further evidence that the impact of these modifications is significant enough to be used). To avoid ambiguity, all results with the original discrete AUC are reported in the supplementary paper.

To ensure a fair comparison, we separated systems trained on INRIA (the majority) from systems trained on TUD-MotionPairs and the only system trained on Caltech in table~\ref{table:all}. For clarity, only systems trained on INRIA were represented in Figure~\ref{fig:all}, however all results for all systems are still reported in table~\ref{table:all}.

\subsection{Results}

In Figure~\ref{fig:inria}, we plot DET curves, i.e. miss rate versus false positives
per image (FPPI), on the fixed INRIA dataset and rank algorithms along two measures: the error rate at 1 FPPI and the area under curve (AUC) rate in the [0, 1] FPPI range. This graph shows the individual contributions of
unsupervised learning (ConvNet-U) and multi-stage features learning
(ConvNet-F-MS) and their combination (ConvNet-U-MS)
compared to the fully-supervised system without multi-stage features
(ConvNet-F). With 17.1\% error rate, unsupervised learning exhibits the most improvements compared to the baseline ConvNet-F (23.39\%). Multi-stage features without unsupervised learning reach 17.29\% error while their combination yields the competitive error rate of 10.55\%.

\begin{figure*}[htb]
\centering
\mbox{\subfigure{\includegraphics[width=.5\linewidth]{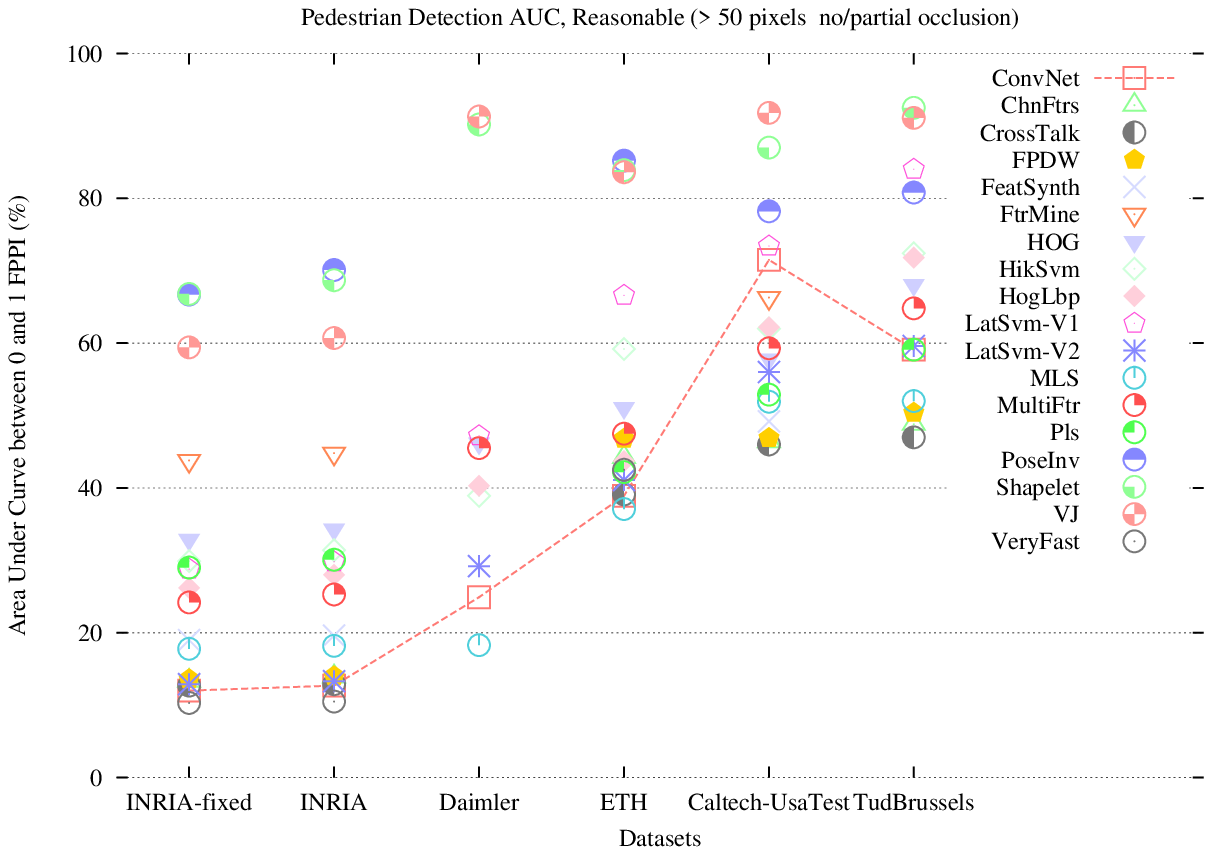}
\quad
\subfigure{\includegraphics[width=.5\linewidth]{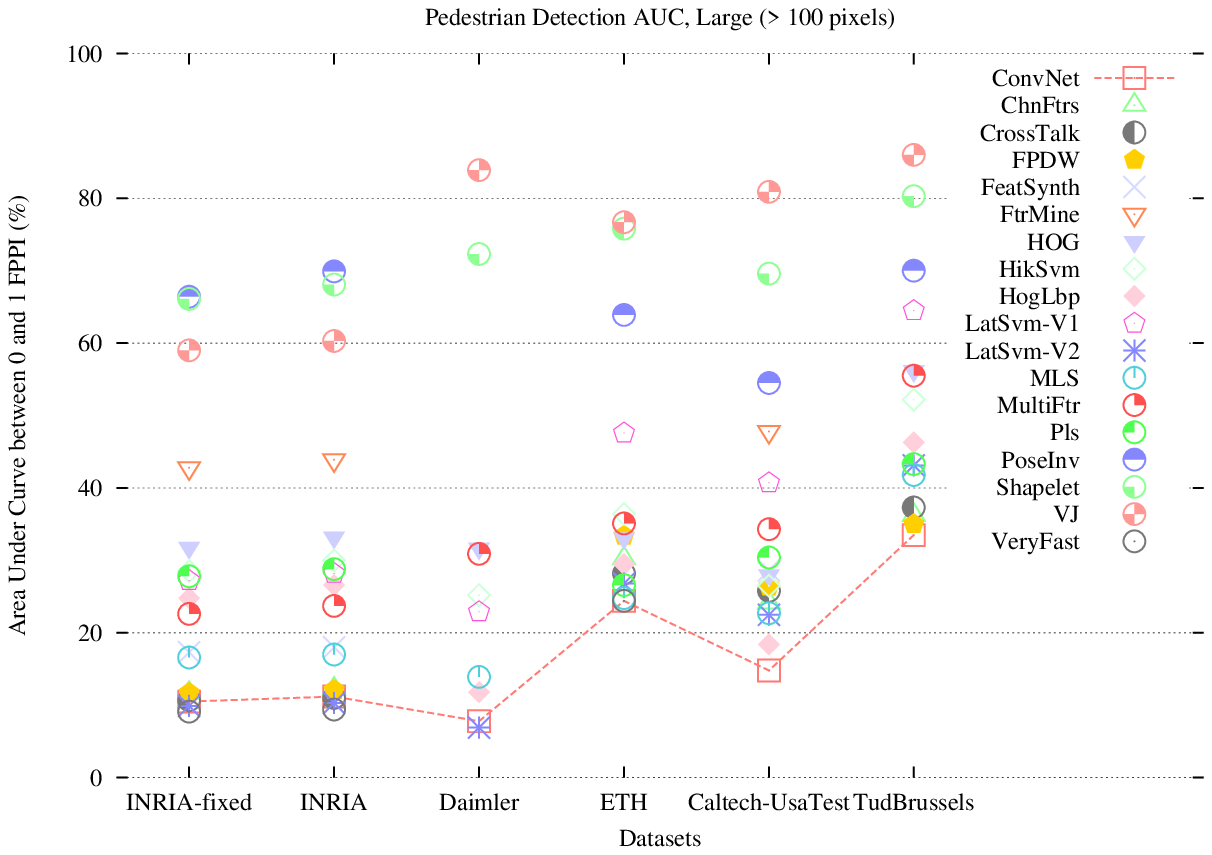} }}}
  

  \caption{\small{\textbf{Reasonable and Large measures for all INRIA-trained systems on all major datasets, using the proposed continuous AUC percentage.} The AUC is computed from DET curves (smaller AUC means more accuracy and less false positives). For a clearer overall performance, each ConvNet point is connected by dotted lines. While only the 'reasonable' and 'large' measures are plotted here, all measures are reported in table~\ref{table:all}. The ConvNet system yields state-of-the-art or competitive results on most datasets and measures, except for the low resolutions measures on the Caltech dataset because of higher reliance on high-resolution cues than other methods.}}
  \label{fig:all}
\end{figure*}

Extensive results comparison of all major pedestrian datasets and published systems is provided in Table~\ref{table:all}. Multiple types of measures proposed by~\cite{dollarCVPR09peds} are reported. For clarity, we also plot in Figure~\ref{fig:all} two of these measures, 'reasonable' and 'large', for INRIA-trained systems. The 'large' plot shows that the ConvNet results in state-of-the-art performance with some margin on the ETH, Caltech and TudBrussels datasets and is closely behind LatSvm-V2 and VeryFast for INRIA and Daimler datasets. In the 'reasonable' plot, the ConvNet yields competitive results for INRIA, Daimler and ETH datasets but performs poorly on the Caltech dataset. 
We suspect the ConvNet with multi-stage features trained at high-resolution is more sensitive to resolution loss than other methods. 
In future work, a ConvNet trained at multiple resolution will likely learn to use appropriate cues for each resolution regime.
 
\newcommand{\sz}{\fontsize{6}{7}\selectfont}

\begin{table*}[htb] 
\sz{
    \begin{center}
      \setlength{\tabcolsep}{.3em}
      \begin{tabular}{ c | c c c c c c c c c c c c c c c c c c}
        \multicolumn{1}{c}{Trained on} & \multicolumn{18}{c}{INRIA} \\
        \hline

        & \textbf{ConvNet} & \textbf{ChnFtrs} & \textbf{CrossTalk} & \textbf{FPDW} & \textbf{FeatSynth} & \textbf{FtrMine} & \textbf{HOG} & \textbf{HikSvm} & \textbf{HogLbp} & \textbf{LatSvm-V1} & \textbf{LatSvm-V2} & \textbf{MLS} & \textbf{MultiFtr} & \textbf{Pls} & \textbf{PoseInv} & \textbf{Shapelet} & \textbf{VJ} & \textbf{VeryFast} \\
        
        \hline
        \multicolumn{19}{c}{All - AUC \%}\\
        \hline
INRIA-fixed & 12.0 & 13.3 & 12.7 & 13.6 & 19.0 & 43.8 & 32.9 & 29.9 & 26.2 & 28.8 & 12.9 & 17.8 & 24.2 & 29.0 & 66.7 & 66.8 & 59.4 & \textbf{10.3} \\
INRIA & 12.7 & 13.9 & 12.9 & 14.0 & 19.6 & 44.8 & 34.3 & 31.4 & 28.0 & 29.8 & 13.3 & 18.2 & 25.3 & 30.1 & 70.1 & 68.7 & 60.7 & \textbf{10.5} \\
Daimler & 58.6 & - & - & - & - & - & 67.9 & 62.4 & 69.8 & 64.2 & 62.3 & \textbf{51.8} & 68.8 & - & - & 94.9 & 94.8 & - \\
ETH & 47.1 & 48.7 & 43.8 & 51.5 & - & - & 54.9 & 61.6 & 51.1 & 69.1 & 49.3 & \textbf{42.8} & 51.7 & 47.4 & 86.5 & 85.6 & 84.5 & 46.9 \\
Caltech-UsaTest & 90.9 & \textbf{77.1} & 77.8 & 78.1 & 78.1 & 86.7 & 85.5 & 86.8 & 87.9 & 91.7 & 84.2 & 83.4 & 83.4 & 81.2 & 92.6 & 95.4 & 99.1 & - \\
TudBrussels & 66.8 & 57.6 & \textbf{55.0} & 59.0 & - & - & 73.6 & 76.4 & 77.2 & 85.7 & 67.2 & 59.2 & 70.5 & 66.1 & 83.8 & 93.8 & 92.7 & - \\

        \multicolumn{19}{c}{Reasonable - AUC \% - \textgreater 50 pixels \& no/partial occlusion}\\
        \hline

INRIA-fixed & 12.0 & 13.3 & 12.7 & 13.6 & 19.0 & 43.8 & 32.9 & 29.9 & 26.2 & 28.8 & 12.9 & 17.8 & 24.2 & 29.0 & 66.7 & 66.8 & 59.4 & \textbf{10.3} \\
INRIA & 12.7 & 13.9 & 12.9 & 14.0 & 19.6 & 44.8 & 34.3 & 31.4 & 28.0 & 29.8 & 13.3 & 18.2 & 25.3 & 30.1 & 70.1 & 68.7 & 60.7 & \textbf{10.5} \\
Daimler & 24.9 & - & - & - & - & - & 46.2 & 38.9 & 40.3 & 47.2 & 29.2 & \textbf{18.3} & 45.5 & - & - & 90.2 & 91.3 & - \\
ETH & 38.9 & 44.2 & 39.1 & 46.8 & - & - & 51.1 & 59.2 & 43.7 & 66.6 & 41.1 & \textbf{37.1} & 47.5 & 42.2 & 85.2 & 83.9 & 83.6 & 42.5 \\
Caltech-UsaTest & 71.5 & 46.4 & \textbf{46.0} & 46.9 & 49.2 & 66.3 & 57.8 & 62.0 & 62.2 & 73.4 & 56.0 & 51.9 & 59.3 & 52.9 & 78.2 & 87.0 & 91.8 & - \\
TudBrussels & 59.1 & 48.8 & \textbf{47.0} & 50.4 & - & - & 68.1 & 72.4 & 71.8 & 84.0 & 59.6 & 52.0 & 64.8 & 59.1 & 80.8 & 92.5 & 91.1 & - \\

\multicolumn{19}{c}{Large - AUC \% - \textgreater 100 pixels}\\
        \hline
INRIA-fixed & 10.5 & 11.6 & 10.7 & 11.7 & 17.3 & 42.8 & 31.8 & 28.5 & 24.8 & 27.2 & 9.9 & 16.6 & 22.6 & 27.8 & 66.4 & 66.1 & 59.0 & \textbf{9.1}\\
INRIA & 11.2 & 12.2 & 11.0 & 12.1 & 18.0 & 43.9 & 33.2 & 30.0 & 26.6 & 28.2 & 10.3 & 17.0 & 23.7 & 28.8 & 69.9 & 68.1 & 60.3 & \textbf{9.4} \\
Daimler & 7.8 & - & - & - & - & - & 31.7 & 25.2 & 11.8 & 22.9 & \textbf{6.9} & 13.9 & 30.9 & - & - & 72.3 & 83.9 & - \\
ETH & \textbf{24.4} & 30.2 & 28.2 & 33.4 & - & - & 33.1 & 36.4 & 29.5 & 47.6 & 26.8 & 24.8 & 35.1 & 26.6 & 63.9 & 75.8 & 76.7 & 24.4 \\
Caltech-UsaTest & \textbf{14.8} & 24.1 & 25.8 & 26.4 & 28.6 & 47.8 & 28.0 & 26.5 & 18.4 & 40.7 & 22.5 & 22.7 & 34.3 & 30.4 & 54.5 & 69.6 & 80.9 & - \\
TudBrussels & \textbf{33.5} & 36.2 & 37.3 & 35.0 & - & - & 56.2 & 52.2 & 46.3 & 64.5 & 43.1 & 41.8 & 55.5 & 43.3 & 70.0 & 80.3 & 86.0 & - \\

        \multicolumn{19}{c}{Near - AUC \% - \textgreater 80 pixels}\\
        \hline
INRIA-fixed & 11.3 & 11.6 & 11.0 & 11.9 & 17.3 & 42.6 & 31.5 & 28.5 & 24.7 & 27.5 & 11.1 & 16.5 & 22.7 & 27.7 & 66.0 & 66.1 & 58.7 & \textbf{9.7} \\
INRIA & 11.9 & 12.2 & 11.2 & 12.3 & 17.9 & 43.7 & 32.9 & 30.0 & 26.5 & 28.5 & 11.5 & 16.8 & 23.8 & 28.8 & 69.4 & 68.1 & 60.0 & \textbf{9.9}\\
Daimler & \textbf{10.0} & - & - & - & - & - & 36.8 & 30.4 & 10.9 & 27.6 & 10.8 & 14.7 & 33.7 & - & - & 78.3 & 86.3 & - \\
ETH & \textbf{28.9} & 35.2 & 30.9 & 37.5 & - & - & 40.5 & 45.6 & 31.7 & 52.2 & 31.4 & 29.5 & 39.4 & 34.1 & 80.6 & 79.9 & 80.0 & 29.8 \\
Caltech-UsaTest & 27.3 & 27.4 & 28.9 & 28.4 & 29.5 & 48.9 & 33.1 & 34.3 & \textbf{24.7} & 47.2 & 26.7 & 29.1 & 40.8 & 31.2 & 66.8 & 75.7 & 85.3 & - \\
TudBrussels & 40.4 & 39.5 & 40.3 & \textbf{38.8} & - & - & 61.1 & 58.7 & 50.5 & 70.9 & 47.1 & 45.3 & 57.2 & 49.6 & 80.0 & 85.6 & 89.0 & - \\
        \multicolumn{19}{c}{Medium - AUC \% - 30-80 pixels}\\
        \hline
INRIA-fixed & 33.1 & 100.0 & 99.7 & 100.0 & 100.0 & 100.0 & 100.0 & 100.0 & 85.3 & 85.3 & 99.7 & 100.0 & 86.1 & 100.0 & 99.7 & 99.7 & 91.5 & \textbf{27.9} \\
INRIA & 33.1 & 100.0 & 99.7 & 100.0 & 100.0 & 100.0 & 100.0 & 100.0 & 85.3 & 85.3 & 99.7 & 100.0 & 86.1 & 100.0 & 99.7 & 99.7 & 91.5 & \textbf{27.9} \\
Daimler & 54.2 & - & - & - & - & - & 62.1 & 54.4 & 70.7 & 58.5 & 60.0 & \textbf{44.7} & 63.2 & - & - & 95.2 & 93.7 & - \\
ETH & 55.4 & 42.9 & \textbf{42.1} & 45.4 & - & - & 49.9 & 54.7 & 61.2 & 71.5 & 57.3 & 43.9 & 47.3 & 45.0 & 73.9 & 74.5 & 71.2 & 48.3 \\
Caltech-UsaTest & 92.2 & \textbf{69.5} & 70.6 & 70.6 & 70.2 & 82.1 & 81.4 & 82.6 & 91.5 & 91.1 & 80.8 & 80.6 & 77.8 & 75.8 & 88.8 & 94.7 & 98.7 & - \\
TudBrussels & 67.8 & 57.4 & \textbf{55.5} & 59.7 & - & - & 71.4 & 74.9 & 82.9 & 85.5 & 68.2 & 59.1 & 68.7 & 65.0 & 79.4 & 94.1 & 91.7 & - \\

      \end{tabular}
    \end{center}
    \caption{\small{\textbf{Performance of all systems on all datasets using the proposed continuous AUC percentage} over the range [0,1] from DET curves. The top performing results (among INRIA-trained models) are highlighted in bold for each row. DET curves plot false positives per image (FPPI) against miss rate. Hence a smaller AUC\% means a more accurate system with lower amount of false positives. The ConvNet model (ConvNet-U-MS here) holds several state-of-the-art or competitive scores. 
We report the multiple measures introduced by~\cite{dollarCVPR09peds} for all major pedestrian datasets. For readibility, not all measures are reported nor are models not trained on INRIA. All results however are reported in the supplementary paper.}
}
    \label{table:all}
  }
\end{table*}

%% file: section_conclusion.tex
\section{Discussion}

We have introduced a new feature learning model with an application to
pedestrian detection. Contrary to popular models where the low-level
features are hand-designed, our model learns all the features at all
levels in a hierarchy. We used the method of~\cite{koray-nips-10} as a
baseline, and extended it by combining high and low resolution
features in the model, and by learning features on the color channels
of the input. Using the INRIA dataset, we have shown that these
improvements provide clear performance benefits. The resulting model
provides state of the art or competitive results on most measures of all publicly available datasets. Small-scale pedestrian measures can be improved in future work by training multiple scale models relying less on high-resolution details.
While computational speed was not the focus and hence was not reported here, our model was successfully used with near real-time speed in a haptic belt system~\cite{haptic}
using parallel hardware. In future work,
models designed for speed combined to highly optimized parallel computing on graphics cards
is expected to yield competitive computational performance.

%% file: supplementary.tex
\section{Evidence for using the proposed continuous Area Under Curve measure}

\begin{figure*}[h]
  \mbox{
    \subfigure[Continuous Area Under Curve\label{auc1_full}]{\includegraphics[width=.5\linewidth]{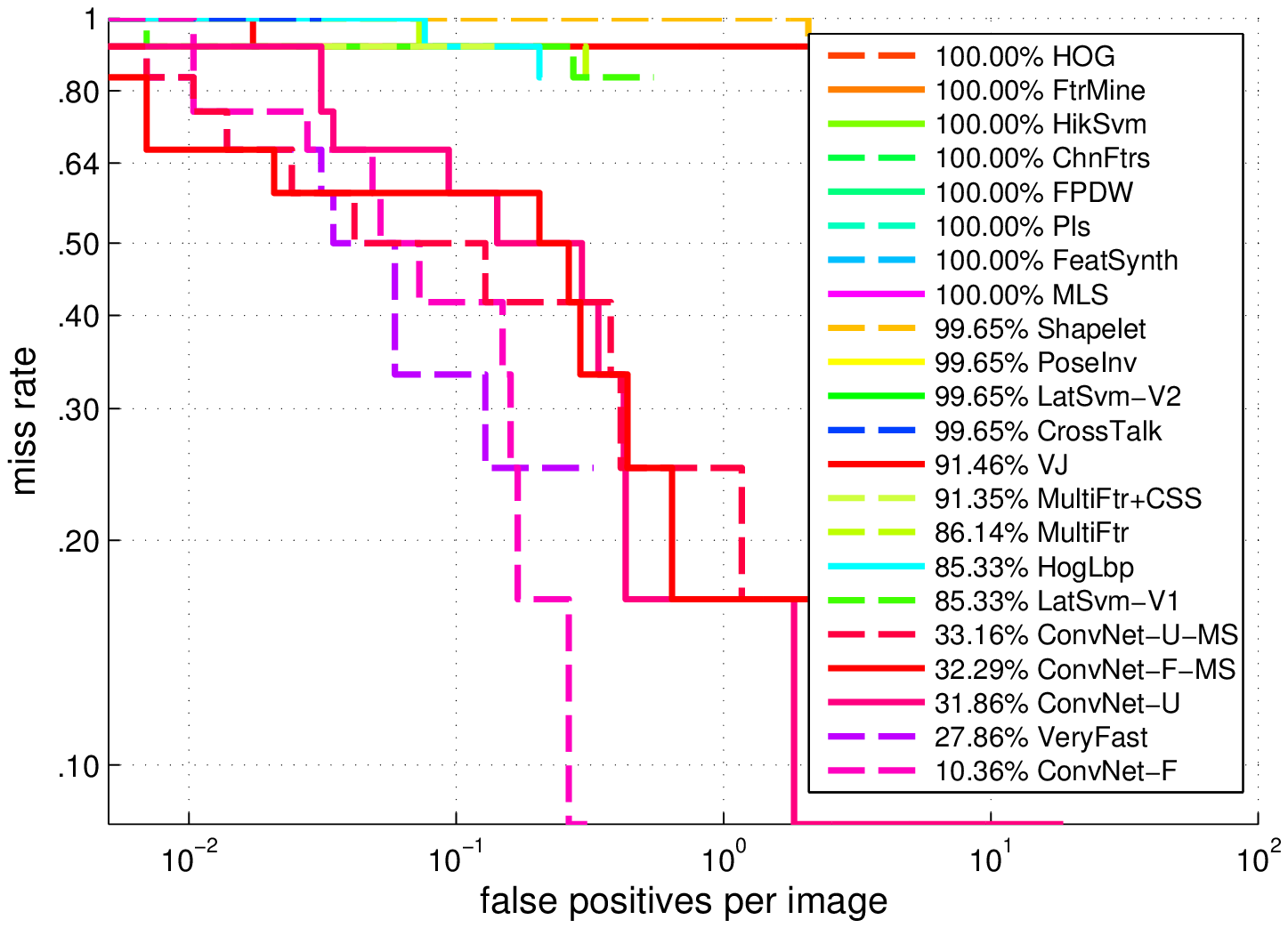}}
    \subfigure[Discrete Area Under Curve\label{auc1_partial}]{\includegraphics[width=.5\linewidth]{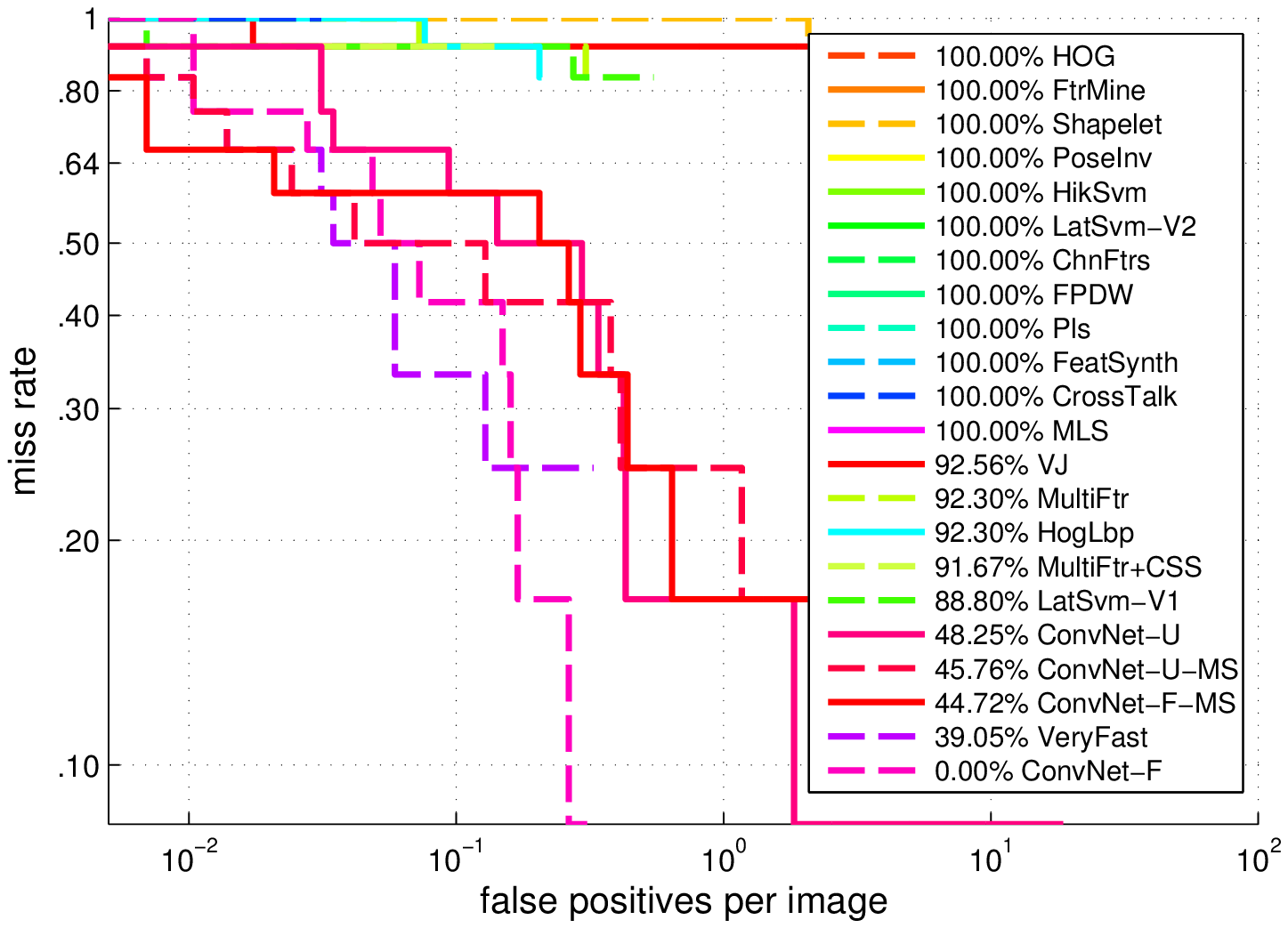}}
  }
  \begin{center}
    \caption{\small{\textbf{Differences between continuous (proposed) and discrete Area Under Curve (AUC) on the INRIA medium scale experiment.}
        In~\ref{auc1_full}, we compute the continuous AUC as opposed to a discrete AUC~\ref{auc1_partial} based on a few points in the standard benchmarking software.~\ref{auc1_partial}
        clearly shows the shortcomings of the discrete AUC which wrongly attributes a 0\% AUC to ConvNet-F instead of 10.36\%.
        Additionally, several models are re-ranked when using the continuous AUC.
        }}
    \label{fig:inria}
  \end{center}
\end{figure*}

\begin{figure*}[h]
  \mbox{
    \subfigure[Continuous Area Under Curve\label{auc2_full}]{\includegraphics[width=.5\linewidth]{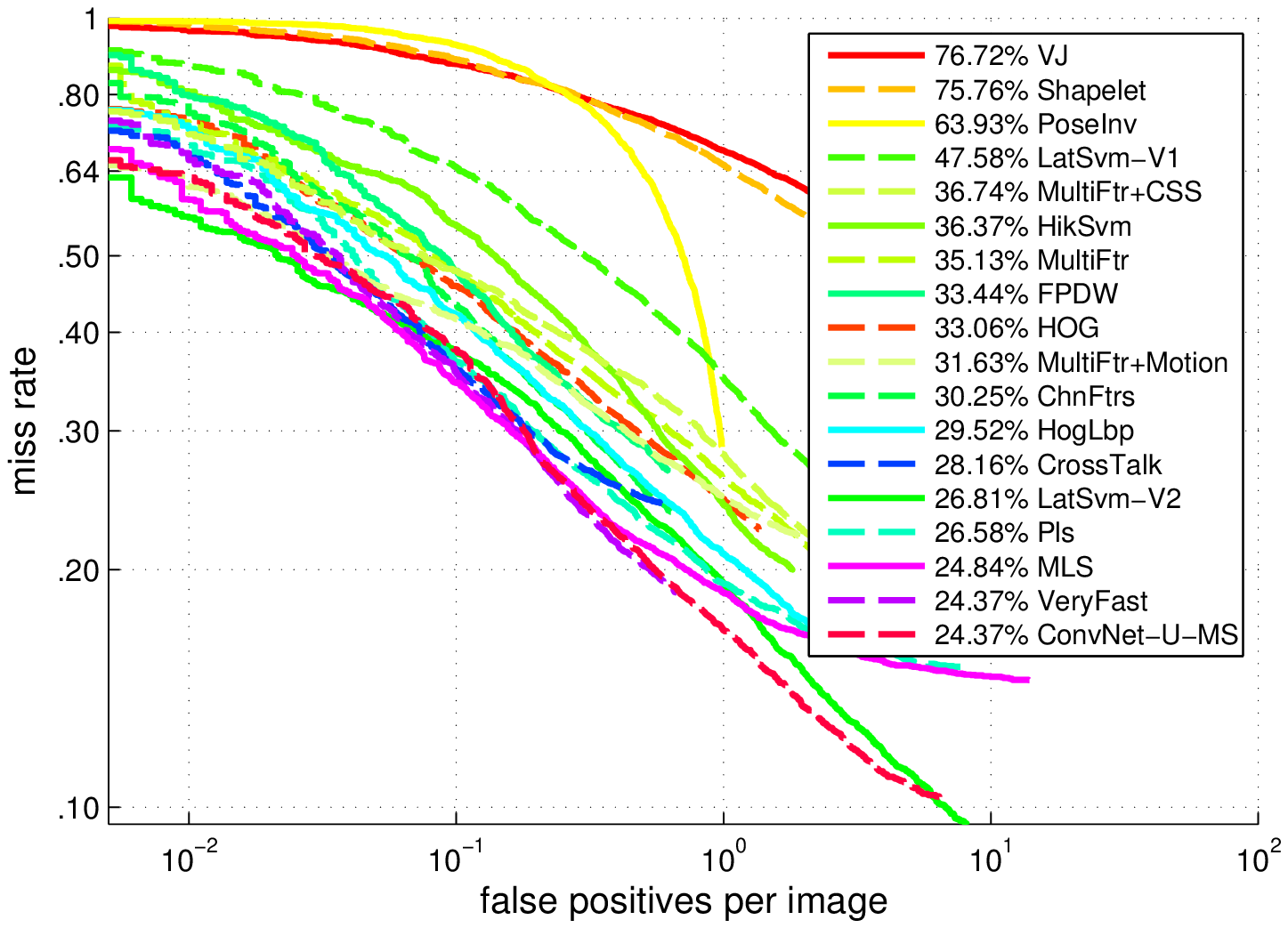}}
    \subfigure[Discrete Area Under Curve\label{auc2_partial}]{\includegraphics[width=.5\linewidth]{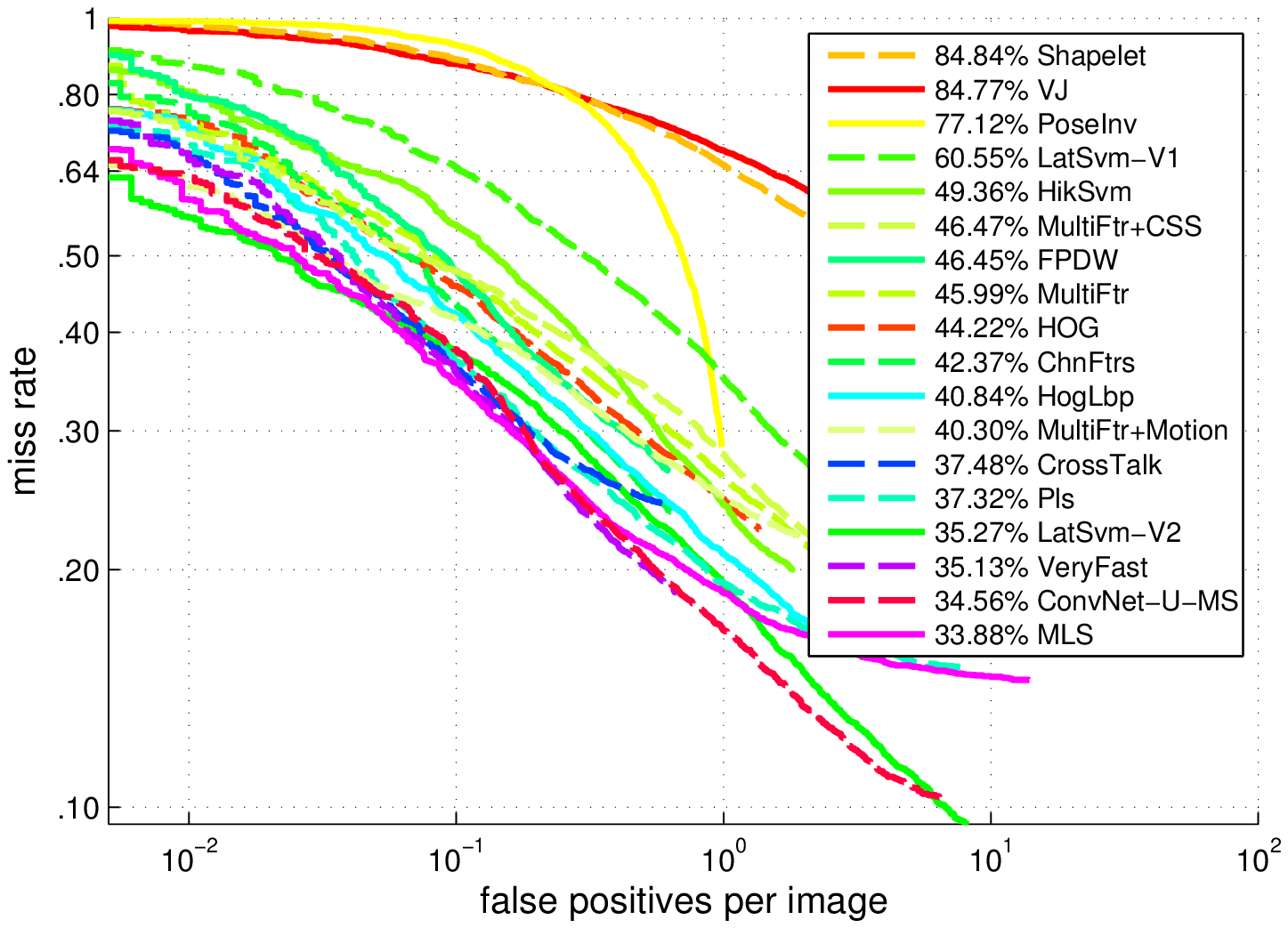}}
  }
  \begin{center}
    \caption{\small{\textbf{Another example of the effects of using the continuous AUC on the ETH large scale experiment.}
        Here several models get re-ranked, including Convnet-U-MS, VeryFast, LatSvm-V2, MultiFtr+Motion, FPDW, MultiFtr+CSS and Shapelet.
    }}
    \label{fig:inria}
  \end{center}
\end{figure*}

\clearpage
\section{Evidence for using the proposed fixed INRIA dataset}
\begin{figure*}[h]
  \mbox{
    \subfigure[Fixed INRIA dataset + Continuous AUC\label{fixed1}]{\includegraphics[width=.5\linewidth]{\inriadir InriaTest-Fixed_roc_exp=scale=large_fullAUC}}
    \subfigure[Original INRIA dataset + Continuous AUC\label{fixed2}]{\includegraphics[width=.5\linewidth]{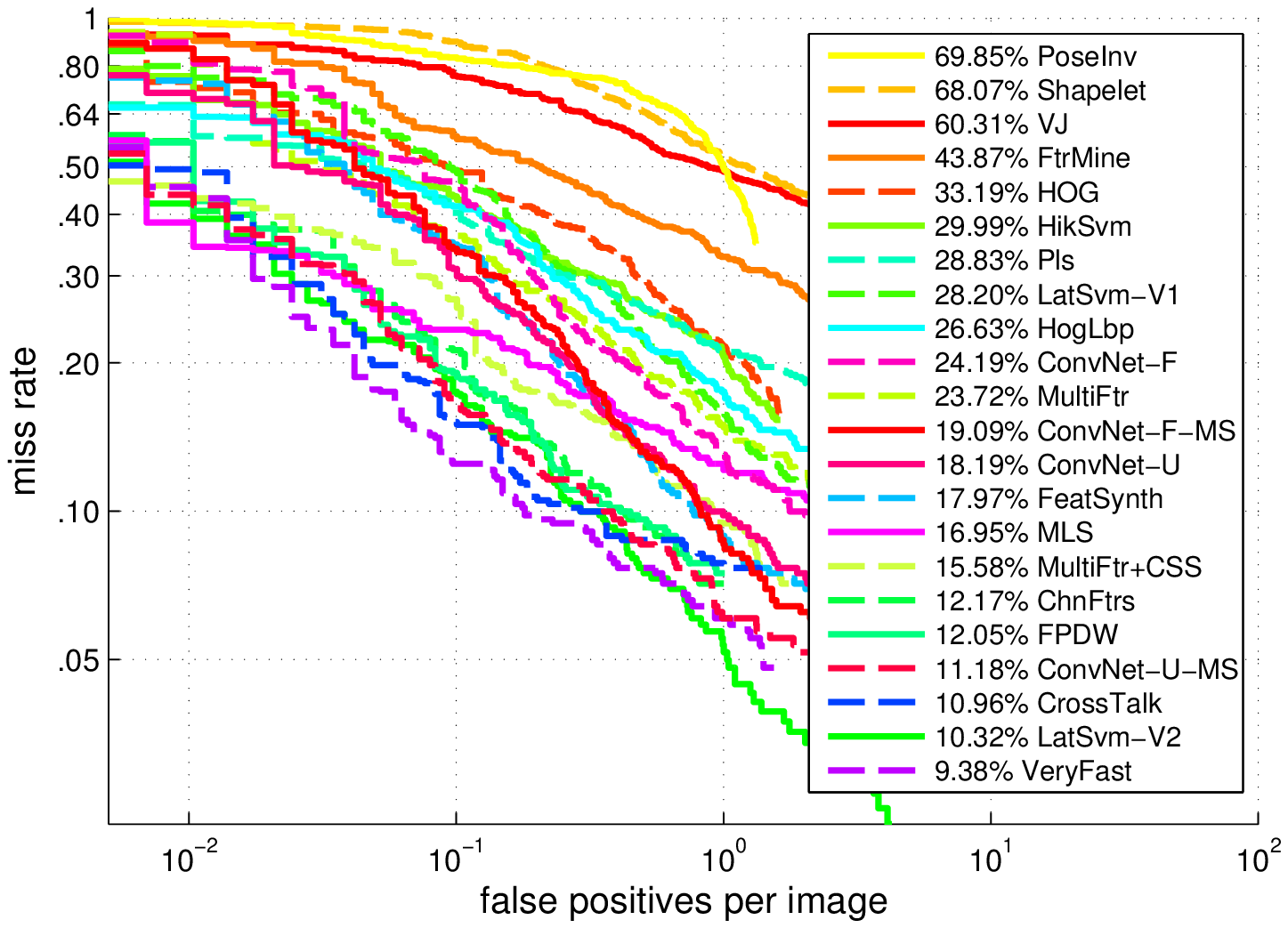}}
  }
  \\
  \mbox{
    \subfigure[Fixed INRIA dataset + Discrete AUC\label{fixed3}]{\includegraphics[width=.5\linewidth]{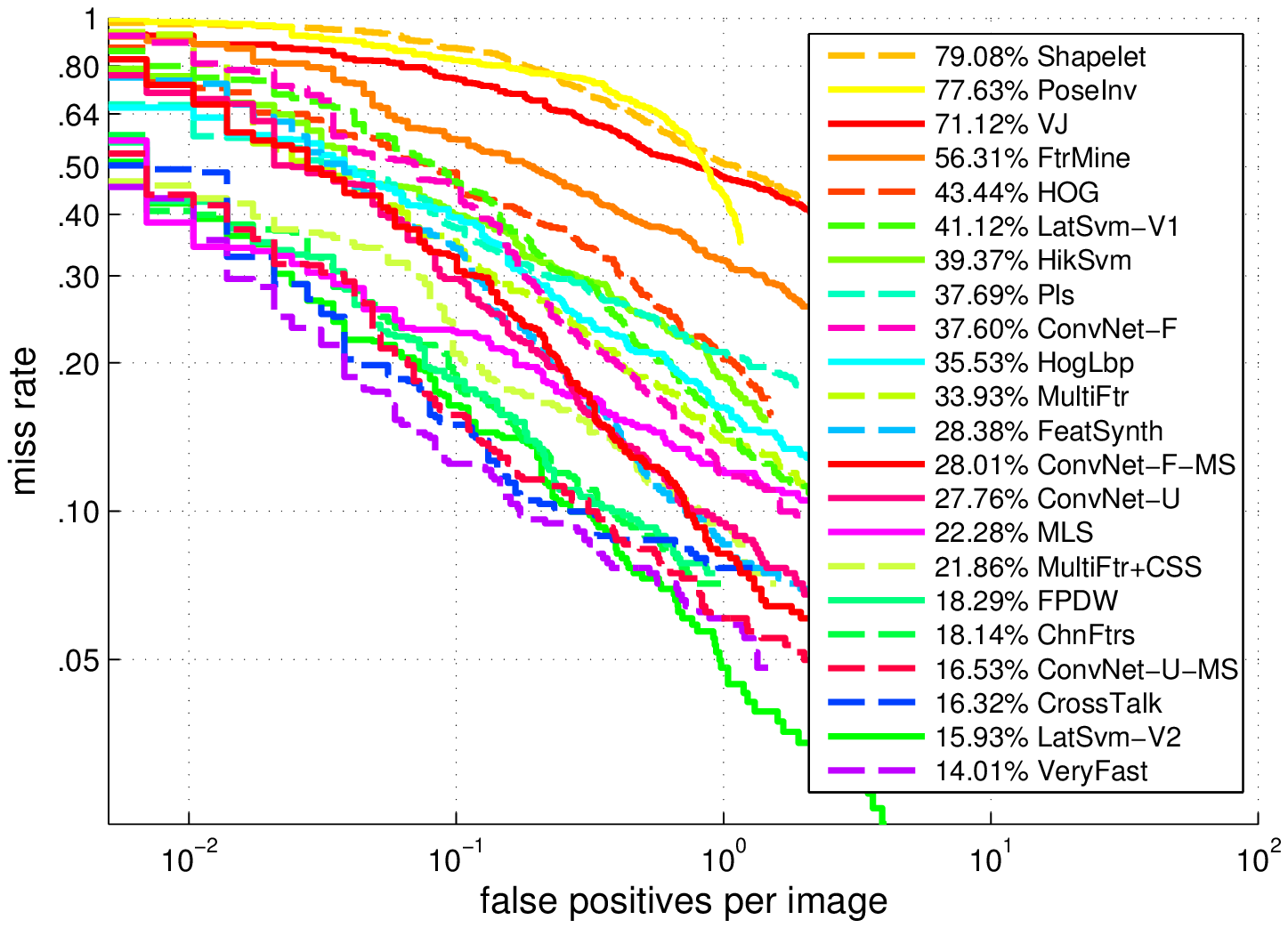}}
    \subfigure[Original INRIA dataset + Discrete AUC\label{fixed4}]{\includegraphics[width=.5\linewidth]{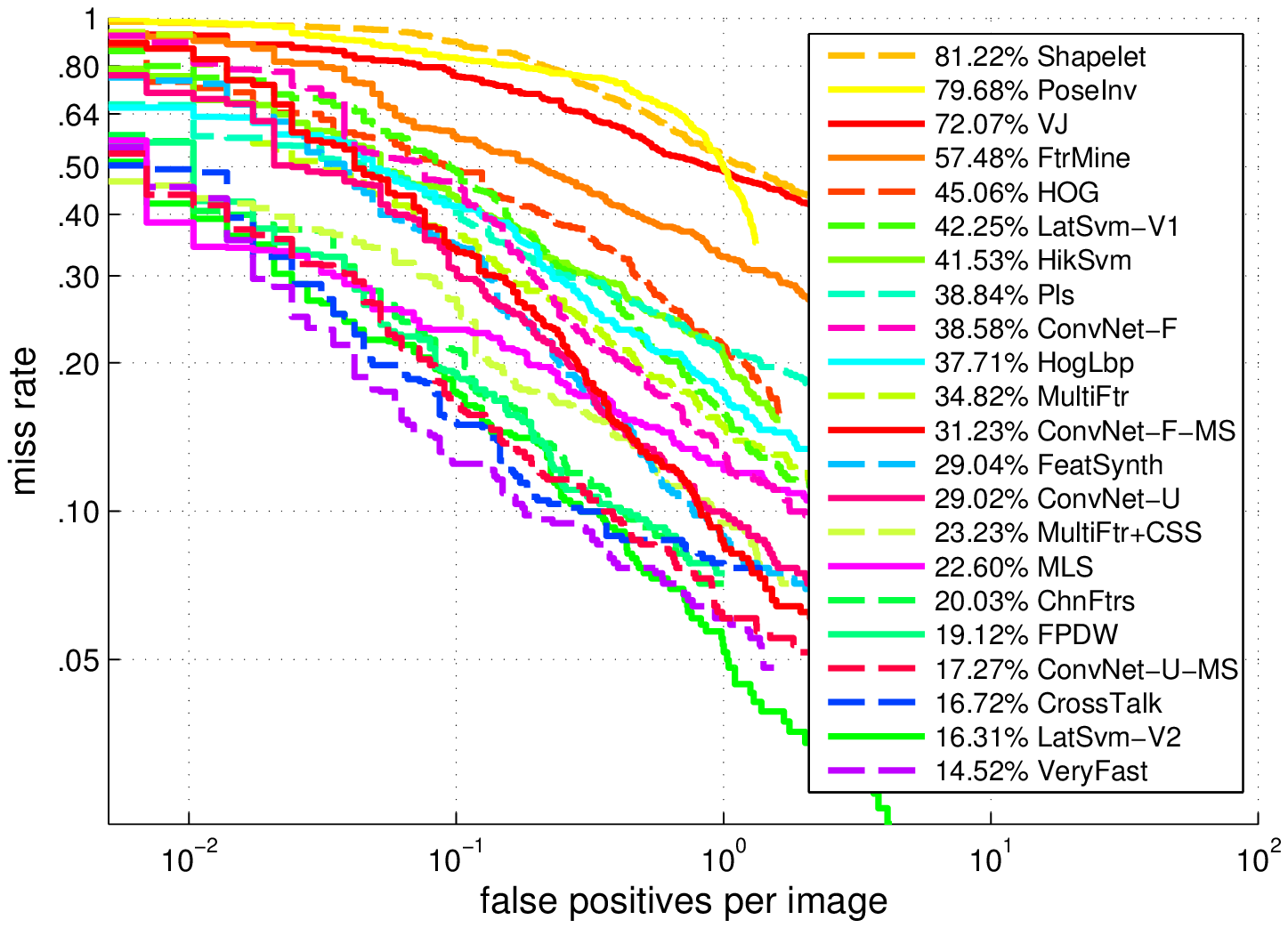}}
  }
  \begin{center}
    \caption{\small{
        \textbf{Effects of fixing INRIA dataset and AUC computation on results.}
        In~\ref{fixed4}, we report the original results obtained with the INRIA dataset and the Area Under Curve (AUC) as computed by the benchmarking software available
        at \urlstyle{same}\url{http://www.vision.caltech.edu/Image\_Datasets/CaltechPedestrians}.
        In~\ref{fixed3}, we use the fixed INRIA dataset instead of the original and observe the re-ranking of several algorithms (ChnFtrs, MLS and ConvNet-F-MS advance by one rank).
        In~\ref{fixed2}, we use the continuous AUC computation instead of the discrete one and observe the following re-ranking: MultiFtr+CSS, FeatSynth, Convnet-F advance by one rank while LatSvm-V1 by two.
        In~\ref{fixed1}, we use both the fixed INRIA and the continuous AUC and observe the following re-ranking as opposed to the unmodified~\ref{fixed4} ranks:
        ConvNet-U-MS, FPDW, MLS, FeathSynth, Convnet-F, and Shapelet advance by one while LatSvm-V1 by two.
        \textbf{Note:} the fixed INRIA dataset and the modified benchmark software are available at \urlstyle{same}\url{http://cs.nyu.edu/~sermanet/data.html}.
        }}
    \label{fig:inria}
  \end{center}
\end{figure*}


\begin{landscape}

\section{All results with the continuous AUC measure}

\begin{table}[h]
\begin{center}
\tiny{
\begin{center}
\setlength{\tabcolsep}{.3em}
\begin{tabular}{ c | c c c c c c c c c c c c c c c c c c c c c | c c | c }
\multicolumn{1}{c}{Trained on} & \multicolumn{21}{c}{INRIA} &\multicolumn{2}{|c}{TUD-MotionPairs} & \multicolumn{1}{|c}{Caltech}\\
\hline
& \textbf{ChnFtrs}& \textbf{ConvNet-F}& \textbf{ConvNet-F-MS}& \textbf{ConvNet-U}& \textbf{ConvNet-U-MS}& \textbf{CrossTalk}& \textbf{FPDW}& \textbf{FeatSynth}& \textbf{FtrMine}& \textbf{HOG}& \textbf{HikSvm}& \textbf{HogLbp}& \textbf{LatSvm-V1}& \textbf{LatSvm-V2}& \textbf{MLS}& \textbf{MultiFtr}& \textbf{Pls}& \textbf{PoseInv}& \textbf{Shapelet}& \textbf{VJ}& \textbf{VeryFast}& \textbf{MultiFtr+CSS}& \textbf{MultiFtr+Motion}& \textbf{MultiResC}\\
\hline
\multicolumn{23}{c}{all - Continuous AUC\%}\\
\hline
INRIA-fixed & 13.3 & 26.7 & 19.5 & 19.8 & 12.0 & 12.7 & 13.6 & 19.0 & 43.8 & 32.9 & 29.9 & 26.2 & 28.8 & 12.9 & 17.8 & 24.2 & 29.0 & 66.7 & 66.8 & 59.4 & \textbf{10.3} & 16.1 & - & -\\
INRIA & 13.9 & 27.4 & 21.4 & 20.9 & 12.7 & 12.9 & 14.0 & 19.6 & 44.8 & 34.3 & 31.4 & 28.0 & 29.8 & 13.3 & 18.2 & 25.3 & 30.1 & 70.1 & 68.7 & 60.7 & \textbf{10.5} & 16.9 & - & -\\
Daimler & - & - & - & - & 58.6 & - & - & - & - & 67.9 & 62.4 & 69.8 & 64.2 & 62.3 & \textbf{51.8} & 68.8 & - & - & 94.9 & 94.8 & - & 48.6 & 40.5 & -\\
ETH & 48.7 & - & - & - & 47.1 & 43.8 & 51.5 & - & - & 54.9 & 61.6 & 51.1 & 69.1 & 49.3 & \textbf{42.8} & 51.7 & 47.4 & 86.5 & 85.6 & 84.5 & 46.9 & 59.5 & 58.2 & -\\
Caltech-UsaTest & \textbf{77.1} & - & - & - & 90.9 & 77.8 & 78.1 & 78.1 & 86.7 & 85.5 & 86.8 & 87.9 & 91.7 & 84.2 & 83.4 & 83.4 & 81.2 & 92.6 & 95.4 & 99.1 & - & 81.2 & 77.9 & 74.2\\
TudBrussels & 57.6 & - & - & - & 66.8 & \textbf{55.0} & 59.0 & - & - & 73.6 & 76.4 & 77.2 & 85.7 & 67.2 & 59.2 & 70.5 & 66.1 & 83.8 & 93.8 & 92.7 & - & 57.8 & 51.4 & -\\

\\
\hline
\multicolumn{23}{c}{reasonable - Continuous AUC\%}\\
\hline
INRIA-fixed & 13.3 & 26.7 & 19.5 & 19.8 & 12.0 & 12.7 & 13.6 & 19.0 & 43.8 & 32.9 & 29.9 & 26.2 & 28.8 & 12.9 & 17.8 & 24.2 & 29.0 & 66.7 & 66.8 & 59.4 & \textbf{10.3} & 16.1 & - & -\\
INRIA & 13.9 & 27.4 & 21.4 & 20.9 & 12.7 & 12.9 & 14.0 & 19.6 & 44.8 & 34.3 & 31.4 & 28.0 & 29.8 & 13.3 & 18.2 & 25.3 & 30.1 & 70.1 & 68.7 & 60.7 & \textbf{10.5} & 16.9 & - & -\\
Daimler & - & - & - & - & 24.9 & - & - & - & - & 46.2 & 38.9 & 40.3 & 47.2 & 29.2 & \textbf{18.3} & 45.5 & - & - & 90.2 & 91.3 & - & 29.0 & 23.3 & -\\
ETH & 44.2 & - & - & - & 38.9 & 39.1 & 46.8 & - & - & 51.1 & 59.2 & 43.7 & 66.6 & 41.1 & \textbf{37.1} & 47.5 & 42.2 & 85.2 & 83.9 & 83.6 & 42.5 & 49.4 & 47.9 & -\\
Caltech-UsaTest & 46.4 & - & - & - & 71.5 & \textbf{46.0} & 46.9 & 49.2 & 66.3 & 57.8 & 62.0 & 62.2 & 73.4 & 56.0 & 51.9 & 59.3 & 52.9 & 78.2 & 87.0 & 91.8 & - & 52.1 & 42.3 & 38.1\\
TudBrussels & 48.8 & - & - & - & 59.1 & \textbf{47.0} & 50.4 & - & - & 68.1 & 72.4 & 71.8 & 84.0 & 59.6 & 52.0 & 64.8 & 59.1 & 80.8 & 92.5 & 91.1 & - & 50.2 & 43.8 & -\\

\\
\hline
\multicolumn{23}{c}{scale=large - Continuous AUC\%}\\
\hline
INRIA-fixed & 11.6 & 23.4 & 17.3 & 17.1 & 10.5 & 10.7 & 11.7 & 17.3 & 42.8 & 31.8 & 28.5 & 24.8 & 27.2 & 9.9 & 16.6 & 22.6 & 27.8 & 66.4 & 66.1 & 59.0 & \textbf{9.1} & 14.8 & - & -\\
INRIA & 12.2 & 24.2 & 19.1 & 18.2 & 11.2 & 11.0 & 12.1 & 18.0 & 43.9 & 33.2 & 30.0 & 26.6 & 28.2 & 10.3 & 17.0 & 23.7 & 28.8 & 69.9 & 68.1 & 60.3 & \textbf{9.4} & 15.6 & - & -\\
Daimler & - & - & - & - & 7.8 & - & - & - & - & 31.7 & 25.2 & 11.8 & 22.9 & \textbf{6.9} & 13.9 & 30.9 & - & - & 72.3 & 83.9 & - & 18.7 & 18.6 & -\\
ETH & 30.2 & - & - & - & \textbf{24.4} & 28.2 & 33.4 & - & - & 33.1 & 36.4 & 29.5 & 47.6 & 26.8 & 24.8 & 35.1 & 26.6 & 63.9 & 75.8 & 76.7 & 24.4 & 36.7 & 31.6 & -\\
Caltech-UsaTest & 24.1 & - & - & - & \textbf{14.8} & 25.8 & 26.4 & 28.6 & 47.8 & 28.0 & 26.5 & 18.4 & 40.7 & 22.5 & 22.7 & 34.3 & 30.4 & 54.5 & 69.6 & 80.9 & - & 28.8 & 12.0 & 12.5\\
TudBrussels & 36.2 & - & - & - & \textbf{33.5} & 37.3 & 35.0 & - & - & 56.2 & 52.2 & 46.3 & 64.5 & 43.1 & 41.8 & 55.5 & 43.3 & 70.0 & 80.3 & 86.0 & - & 45.3 & 38.1 & -\\

\\
\hline
\multicolumn{23}{c}{scale=near - Continuous AUC\%}\\
\hline
INRIA-fixed & 11.6 & 24.5 & 18.6 & 17.7 & 11.3 & 11.0 & 11.9 & 17.3 & 42.6 & 31.5 & 28.5 & 24.7 & 27.5 & 11.1 & 16.5 & 22.7 & 27.7 & 66.0 & 66.1 & 58.7 & \textbf{9.7} & 14.7 & - & -\\
INRIA & 12.2 & 25.3 & 20.5 & 18.8 & 11.9 & 11.2 & 12.3 & 17.9 & 43.7 & 32.9 & 30.0 & 26.5 & 28.5 & 11.5 & 16.8 & 23.8 & 28.8 & 69.4 & 68.1 & 60.0 & \textbf{9.9} & 15.5 & - & -\\
Daimler & - & - & - & - & \textbf{10.0} & - & - & - & - & 36.8 & 30.4 & 10.9 & 27.6 & 10.8 & 14.7 & 33.7 & - & - & 78.3 & 86.3 & - & 18.4 & 19.5 & -\\
ETH & 35.2 & - & - & - & \textbf{28.9} & 30.9 & 37.5 & - & - & 40.5 & 45.6 & 31.7 & 52.2 & 31.4 & 29.5 & 39.4 & 34.1 & 80.6 & 79.9 & 80.0 & 29.8 & 40.0 & 36.3 & -\\
Caltech-UsaTest & 27.4 & - & - & - & 27.3 & 28.9 & 28.4 & 29.5 & 48.9 & 33.1 & 34.3 & \textbf{24.7} & 47.2 & 26.7 & 29.1 & 40.8 & 31.2 & 66.8 & 75.7 & 85.3 & - & 30.4 & 16.4 & 15.0\\
TudBrussels & 39.5 & - & - & - & 40.4 & 40.3 & \textbf{38.8} & - & - & 61.1 & 58.7 & 50.5 & 70.9 & 47.1 & 45.3 & 57.2 & 49.6 & 80.0 & 85.6 & 89.0 & - & 46.5 & 39.8 & -\\

\\
\hline
\multicolumn{23}{c}{scale=medium - Continuous AUC\%}\\
\hline
INRIA-fixed & 100.0 & \textbf{10.4} & 32.3 & 31.9 & 33.2 & 99.7 & 100.0 & 100.0 & 100.0 & 100.0 & 100.0 & 85.3 & 85.3 & 99.7 & 100.0 & 86.1 & 100.0 & 99.7 & 99.7 & 91.5 & 27.9 & 91.3 & - & -\\
INRIA & 100.0 & \textbf{10.4} & 32.3 & 31.9 & 33.2 & 99.7 & 100.0 & 100.0 & 100.0 & 100.0 & 100.0 & 85.3 & 85.3 & 99.7 & 100.0 & 86.1 & 100.0 & 99.7 & 99.7 & 91.5 & 27.9 & 91.3 & - & -\\
Daimler & - & - & - & - & 54.2 & - & - & - & - & 62.1 & 54.4 & 70.7 & 58.5 & 60.0 & \textbf{44.7} & 63.2 & - & - & 95.2 & 93.7 & - & 43.4 & 34.0 & -\\
ETH & 42.9 & - & - & - & 55.4 & \textbf{42.1} & 45.4 & - & - & 49.9 & 54.7 & 61.2 & 71.5 & 57.3 & 43.9 & 47.3 & 45.0 & 73.9 & 74.5 & 71.2 & 48.3 & 55.2 & 54.7 & -\\
Caltech-UsaTest & \textbf{69.5} & - & - & - & 92.2 & 70.6 & 70.6 & 70.2 & 82.1 & 81.4 & 82.6 & 91.5 & 91.1 & 80.8 & 80.6 & 77.8 & 75.8 & 88.8 & 94.7 & 98.7 & - & 76.0 & 73.4 & 65.4\\
TudBrussels & 57.4 & - & - & - & 67.8 & \textbf{55.5} & 59.7 & - & - & 71.4 & 74.9 & 82.9 & 85.5 & 68.2 & 59.1 & 68.7 & 65.0 & 79.4 & 94.1 & 91.7 & - & 55.0 & 48.6 & -\\

\\
\hline
\multicolumn{23}{c}{scale=far - Continuous AUC\%}\\
\hline
Caltech-UsaTest & \textbf{93.4} & - & - & - & 100.0 & 95.4 & 94.2 & 94.7 & 95.0 & 96.2 & 98.2 & 100.0 & 97.8 & 97.4 & 100.0 & 95.9 & 98.7 & 100.0 & 99.9 & 99.3 & - & 95.3 & 94.6 & 100.0\\

\\
\hline
\multicolumn{23}{c}{occ=partial - Continuous AUC\%}\\
\hline
Caltech-UsaTest & \textbf{61.3} & - & - & - & 81.9 & 69.2 & 66.9 & 67.0 & 81.6 & 77.1 & 80.3 & 75.0 & 84.1 & 76.7 & 71.3 & 78.6 & 68.1 & 85.7 & 90.5 & 96.9 & - & 76.6 & 64.4 & 61.3\\

\\
\hline
\multicolumn{23}{c}{occ=heavy - Continuous AUC\%}\\
\hline
Caltech-UsaTest & 91.3 & - & - & - & 96.3 & 90.7 & 91.8 & \textbf{87.8} & 95.5 & 93.7 & 93.2 & 95.6 & 94.4 & 93.3 & 92.1 & 95.0 & 92.3 & 97.2 & 97.3 & 98.2 & - & 91.1 & 87.8 & 84.0\\

\\
\hline
\multicolumn{23}{c}{ar=atypical - Continuous AUC\%}\\
\hline
Caltech-UsaTest & 57.3 & - & - & - & 74.5 & \textbf{55.4} & 61.9 & 60.3 & 76.8 & 75.3 & 77.4 & 77.2 & 85.8 & 70.8 & 71.5 & 73.3 & 69.1 & 86.0 & 92.1 & 93.6 & - & 64.3 & 48.9 & 50.6\\

\end{tabular}
\end{center}
}
    \caption{\small{\textbf{Results for all experiments on all datasets using the proposed continuous AUC.} The top performing results (INRIA-trained only) are highlighted in bold for each row.
        The continuous AUC percentage is taken over the range [0,1] from DET curves. DET curves plot false positives per image (FPPI) against miss rate.
        Hence a smaller AUC\% means a more accurate system with greater reduction of false positives.
}
    }
    \label{table:all2}
\end{center}
\end{table}
\end{landscape}


\begin{landscape}

\section{All results with the original discrete AUC measure}

\begin{table}[h]
\begin{center}
\tiny{
\begin{center}
\setlength{\tabcolsep}{.3em}
\begin{tabular}{ c | c c c c c c c c c c c c c c c c c c c c c | c c | c }
\multicolumn{1}{c}{Trained on} & \multicolumn{21}{c}{INRIA} &\multicolumn{2}{|c}{TUD-MotionPairs} & \multicolumn{1}{|c}{Caltech}\\
\hline
& \textbf{ChnFtrs}& \textbf{ConvNet-F}& \textbf{ConvNet-F-MS}& \textbf{ConvNet-U}& \textbf{ConvNet-U-MS}& \textbf{CrossTalk}& \textbf{FPDW}& \textbf{FeatSynth}& \textbf{FtrMine}& \textbf{HOG}& \textbf{HikSvm}& \textbf{HogLbp}& \textbf{LatSvm-V1}& \textbf{LatSvm-V2}& \textbf{MLS}& \textbf{MultiFtr}& \textbf{Pls}& \textbf{PoseInv}& \textbf{Shapelet}& \textbf{VJ}& \textbf{VeryFast}& \textbf{MultiFtr+CSS}& \textbf{MultiFtr+Motion}& \textbf{MultiResC}\\
\hline
\multicolumn{23}{c}{all - Discrete AUC\%}\\
\hline
INRIA-fixed & 20.2 & 41.7 & 30.5 & 31.5 & 18.7 & 18.6 & 20.6 & 30.3 & 57.2 & 44.4 & 40.7 & 36.9 & 42.9 & 19.6 & 23.6 & 35.6 & 39.0 & 77.9 & 79.6 & 71.6 & \textbf{15.8} & 23.3 & - & -\\
INRIA & 22.2 & 42.2 & 34.2 & 32.6 & 19.8 & 19.0 & 21.5 & 30.9 & 58.3 & 46.0 & 42.8 & 39.1 & 43.8 & 20.0 & 23.9 & 36.5 & 40.1 & 80.1 & 81.7 & 72.5 & \textbf{16.0} & 24.7 & - & -\\
Daimler & - & - & - & - & 64.5 & - & - & - & - & 77.4 & 74.3 & 74.8 & 72.2 & 68.3 & \textbf{60.2} & 76.6 & - & - & 96.8 & 96.7 & - & 58.2 & 48.1 & -\\
ETH & 61.9 & - & - & - & 57.8 & 56.5 & 64.3 & - & - & 67.3 & 73.8 & 61.7 & 78.6 & 58.3 & \textbf{54.9} & 63.6 & 59.8 & 92.8 & 92.0 & 90.4 & 59.3 & 70.7 & 70.1 & -\\
Caltech-UsaTest & 82.8 & - & - & - & 92.9 & \textbf{82.7} & 83.7 & 84.1 & 90.6 & 90.4 & 91.4 & 89.8 & 93.7 & 87.7 & 87.2 & 87.8 & 85.7 & 95.5 & 97.0 & 99.5 & - & 85.5 & 82.8 & 80.4\\
TudBrussels & 67.7 & - & - & - & 74.9 & \textbf{65.1} & 70.1 & - & - & 81.8 & 85.3 & 85.4 & 91.3 & 75.6 & 69.0 & 78.1 & 76.2 & 90.0 & 96.2 & 95.5 & - & 66.5 & 62.0 & -\\

\\
\hline
\multicolumn{23}{c}{reasonable - Discrete AUC\%}\\
\hline
INRIA-fixed & 20.2 & 41.7 & 30.5 & 31.5 & 18.7 & 18.6 & 20.6 & 30.3 & 57.2 & 44.4 & 40.7 & 36.9 & 42.9 & 19.6 & 23.6 & 35.6 & 39.0 & 77.9 & 79.6 & 71.6 & \textbf{15.8} & 23.3 & - & -\\
INRIA & 22.2 & 42.2 & 34.2 & 32.6 & 19.8 & 19.0 & 21.5 & 30.9 & 58.3 & 46.0 & 42.8 & 39.1 & 43.8 & 20.0 & 23.9 & 36.5 & 40.1 & 80.1 & 81.7 & 72.5 & \textbf{16.0} & 24.7 & - & -\\
Daimler & - & - & - & - & 32.5 & - & - & - & - & 59.8 & 55.0 & 48.7 & 57.6 & 38.0 & \textbf{27.6} & 57.2 & - & - & 93.8 & 94.7 & - & 39.2 & 29.2 & -\\
ETH & 57.5 & - & - & - & 50.3 & 51.9 & 60.1 & - & - & 64.2 & 72.0 & 55.2 & 76.7 & 50.9 & \textbf{49.4} & 59.8 & 54.9 & 92.1 & 90.9 & 89.9 & 54.8 & 60.7 & 60.0 & -\\
Caltech-UsaTest & 56.3 & - & - & - & 77.2 & \textbf{53.9} & 57.4 & 60.2 & 74.4 & 68.5 & 73.4 & 67.8 & 79.8 & 63.3 & 61.0 & 68.3 & 62.1 & 86.3 & 91.4 & 94.7 & - & 60.9 & 50.9 & 48.5\\
TudBrussels & 60.3 & - & - & - & 68.8 & \textbf{58.0} & 63.0 & - & - & 77.9 & 82.5 & 81.7 & 90.2 & 69.6 & 62.6 & 73.4 & 70.7 & 88.0 & 95.4 & 94.5 & - & 59.5 & 54.8 & -\\

\\
\hline
\multicolumn{23}{c}{scale=large - Discrete AUC\%}\\
\hline
INRIA-fixed & 18.1 & 37.6 & 28.0 & 27.8 & 16.5 & 16.3 & 18.3 & 28.4 & 56.3 & 43.4 & 39.4 & 35.5 & 41.1 & 15.9 & 22.3 & 33.9 & 37.7 & 77.6 & 79.1 & 71.1 & \textbf{14.0} & 21.9 & - & -\\
INRIA & 20.0 & 38.6 & 31.2 & 29.0 & 17.3 & 16.7 & 19.1 & 29.0 & 57.5 & 45.1 & 41.5 & 37.7 & 42.2 & 16.3 & 22.6 & 34.8 & 38.8 & 79.7 & 81.2 & 72.1 & \textbf{14.5} & 23.2 & - & -\\
Daimler & - & - & - & - & 11.6 & - & - & - & - & 47.4 & 40.5 & 17.0 & 32.8 & \textbf{10.6} & 20.6 & 43.4 & - & - & 81.6 & 89.6 & - & 23.7 & 21.4 & -\\
ETH & 42.4 & - & - & - & 34.6 & 37.5 & 46.4 & - & - & 44.2 & 49.4 & 40.8 & 60.6 & 35.3 & \textbf{33.9} & 46.0 & 37.3 & 77.1 & 84.8 & 84.8 & 35.1 & 46.5 & 40.3 & -\\
Caltech-UsaTest & 30.2 & - & - & - & \textbf{21.4} & 29.8 & 33.4 & 36.2 & 59.1 & 37.9 & 39.0 & 22.7 & 49.9 & 28.2 & 31.1 & 43.0 & 36.5 & 71.6 & 76.6 & 86.2 & - & 36.3 & 15.7 & 17.8\\
TudBrussels & 43.1 & - & - & - & \textbf{41.1} & 45.2 & 44.1 & - & - & 71.2 & 67.6 & 59.3 & 76.6 & 54.2 & 51.6 & 63.5 & 54.3 & 82.3 & 87.1 & 91.9 & - & 54.9 & 46.5 & -\\

\\
\hline
\multicolumn{23}{c}{scale=near - Discrete AUC\%}\\
\hline
INRIA-fixed & 18.3 & 38.7 & 29.4 & 28.3 & 17.7 & 16.7 & 18.7 & 28.3 & 56.2 & 43.1 & 39.3 & 35.4 & 41.5 & 17.5 & 22.2 & 34.1 & 37.7 & 77.4 & 79.1 & 70.9 & \textbf{14.9} & 21.7 & - & -\\
INRIA & 20.2 & 39.8 & 33.2 & 29.4 & 18.8 & 17.1 & 19.5 & 28.9 & 57.3 & 44.7 & 41.5 & 37.6 & 42.5 & 17.9 & 22.5 & 35.0 & 38.8 & 79.5 & 81.3 & 71.9 & \textbf{15.1} & 23.2 & - & -\\
Daimler & - & - & - & - & \textbf{14.8} & - & - & - & - & 52.0 & 45.5 & 15.6 & 37.9 & 16.1 & 22.5 & 46.2 & - & - & 85.6 & 91.3 & - & 25.2 & 23.6 & -\\
ETH & 48.3 & - & - & - & \textbf{39.3} & 42.1 & 50.7 & - & - & 53.5 & 59.2 & 43.0 & 64.7 & 40.6 & 40.0 & 50.3 & 46.0 & 89.1 & 88.0 & 87.2 & 41.1 & 51.0 & 45.4 & -\\
Caltech-UsaTest & 35.1 & - & - & - & 36.7 & 34.3 & 36.7 & 38.9 & 58.4 & 44.0 & 48.0 & \textbf{30.8} & 57.1 & 34.3 & 37.7 & 50.0 & 39.8 & 77.9 & 82.9 & 89.9 & - & 38.8 & 21.7 & 21.2\\
TudBrussels & \textbf{49.6} & - & - & - & 49.9 & 49.7 & 50.0 & - & - & 73.5 & 73.0 & 64.3 & 81.3 & 58.0 & 55.2 & 66.3 & 61.6 & 88.0 & 90.9 & 93.0 & - & 56.3 & 50.1 & -\\

\\
\hline
\multicolumn{23}{c}{scale=medium - Discrete AUC\%}\\
\hline
INRIA-fixed & 100.0 & \textbf{0.0} & 44.7 & 48.3 & 45.8 & 100.0 & 100.0 & 100.0 & 100.0 & 100.0 & 100.0 & 92.3 & 88.8 & 100.0 & 100.0 & 92.3 & 100.0 & 100.0 & 100.0 & 92.6 & 39.0 & 91.7 & - & -\\
INRIA & 100.0 & \textbf{0.0} & 44.7 & 48.3 & 45.8 & 100.0 & 100.0 & 100.0 & 100.0 & 100.0 & 100.0 & 92.3 & 88.8 & 100.0 & 100.0 & 92.3 & 100.0 & 100.0 & 100.0 & 92.6 & 39.0 & 91.7 & - & -\\
Daimler & - & - & - & - & 60.7 & - & - & - & - & 72.8 & 68.0 & 75.0 & 67.8 & 67.1 & \textbf{53.9} & 72.2 & - & - & 97.0 & 96.1 & - & 53.5 & 41.2 & -\\
ETH & 53.6 & - & - & - & 65.7 & \textbf{53.0} & 56.0 & - & - & 59.8 & 67.6 & 70.3 & 80.3 & 67.1 & 54.2 & 58.0 & 55.9 & 85.0 & 82.6 & 80.6 & 60.5 & 67.5 & 67.8 & -\\
Caltech-UsaTest & \textbf{77.4} & - & - & - & 94.9 & 77.4 & 78.4 & 78.2 & 87.3 & 87.4 & 88.5 & 93.0 & 93.3 & 85.5 & 85.5 & 84.1 & 82.0 & 93.1 & 96.7 & 99.4 & - & 82.1 & 80.1 & 73.2\\
TudBrussels & 67.7 & - & - & - & 76.5 & \textbf{64.9} & 70.6 & - & - & 79.4 & 84.4 & 90.2 & 91.1 & 77.2 & 69.2 & 76.7 & 75.7 & 86.8 & 96.5 & 94.7 & - & 63.8 & 58.1 & -\\

\\
\hline
\multicolumn{23}{c}{scale=far - Discrete AUC\%}\\
\hline
Caltech-UsaTest & \textbf{95.2} & - & - & - & 100.0 & 96.6 & 95.7 & 96.4 & 96.9 & 97.1 & 99.1 & 100.0 & 98.5 & 97.9 & 100.0 & 96.9 & 99.2 & 100.0 & 99.9 & 99.7 & - & 96.7 & 96.8 & 100.0\\

\\
\hline
\multicolumn{23}{c}{occ=partial - Discrete AUC\%}\\
\hline
Caltech-UsaTest & \textbf{73.0} & - & - & - & 87.1 & 76.1 & 77.6 & 75.8 & 88.3 & 84.5 & 88.3 & 79.9 & 89.1 & 81.3 & 79.7 & 85.9 & 75.0 & 92.3 & 93.5 & 98.7 & - & 81.4 & 73.0 & 71.6\\

\\
\hline
\multicolumn{23}{c}{occ=heavy - Discrete AUC\%}\\
\hline
Caltech-UsaTest & 94.8 & - & - & - & 97.6 & 93.7 & 95.6 & \textbf{92.9} & 97.7 & 96.0 & 95.3 & 96.7 & 96.4 & 95.5 & 94.7 & 96.5 & 94.8 & 98.3 & 98.3 & 98.8 & - & 93.7 & 92.6 & 90.3\\

\\
\hline
\multicolumn{23}{c}{ar=atypical - Discrete AUC\%}\\
\hline
Caltech-UsaTest & 67.7 & - & - & - & 81.1 & \textbf{64.3} & 72.4 & 71.7 & 85.6 & 85.1 & 86.2 & 83.4 & 89.7 & 77.4 & 79.5 & 82.4 & 77.8 & 91.9 & 95.0 & 96.7 & - & 74.1 & 57.1 & 61.8\\

\end{tabular}
\end{center}
}
\caption{\small{\textbf{Results for all experiments on all datasets using the discrete AUC.} This table is identical to table~\ref{table:all2} except it is using the discrete AUC instead of the proposed continuous AUC.
    }}
    \label{table:all3}
\end{center}
\end{table}
\end{landscape}